\documentclass[sn-mathphys,iicol]{sn-jnl}

\usepackage{times}
\usepackage{graphicx}
\usepackage{amsmath,amssymb,amsfonts}
\usepackage{textcomp}
\usepackage{xcolor}
\usepackage{algorithm}
\usepackage{subcaption}
\usepackage{color, soul}
\usepackage{booktabs}
\usepackage{balance}
\usepackage{hyperref}

\usepackage{program}
\catcode`\|=12\relax
\usepackage{tikz}
\usetikzlibrary{calc}
\usepackage{mwe}

\newcommand{\etal}{{\textit{et al}}.\@ }

    \usepackage{etoolbox}
    \makeatletter
    \patchcmd{\ps@headings}
    {\hbox to \hsize{\hfill Springer Nature 2021 \LaTeX\ template\hfill}}
    {\hbox to \hsize{\hfill Based on  Springer Nature \LaTeX\ template\hfill}}
    {}
    {}
    \patchcmd{\ps@titlepage}
    {\hbox to \hsize{\hfill Springer Nature 2021 \LaTeX\ template\hfill}}
    {\hbox to \hsize{\hfill Based on  Springer Nature \LaTeX\ template\hfill}}
    {}
    {}
    \makeatother

\jyear{2023}%

\theoremstyle{thmstyleone}%
\theoremstyle{thmstyletwo}%

\theoremstyle{thmstylethree}%

\begin{document}

\title[Routing in Multi-Path Networks]{End-to-end Data-Dependent Routing in Multi-Path Neural Networks}

\author*[1,2]{\fnm{Dumindu} \sur{Tissera}}
\author[1,2]{\fnm{Rukshan} \sur{Wijesinghe}}
\author[2]{\fnm{Kasun} \sur{Vithanage}}
\author[2]{\fnm{Alex} \sur{Xavier}}
\author[2]{\fnm{Subha} \sur{Fernando}}
\author[1,2]{\fnm{Ranga} \sur{Rodrigo}}

\affil*[1]{\orgdiv{Department of Electronics and Telecommunication Engineering}, \orgname{University of Moratuwa}, \orgaddress{\country{Sri Lanka}}}
\affil[2]{\orgdiv{CodeGen QBITS Lab}, \orgname{University of Moratuwa}, \orgaddress{\country{Sri Lanka}}}

\abstract{Neural networks are known to give better performance with increased depth due to their ability to learn more abstract features. Although the deepening of networks has been well established, there is still room for efficient feature extraction within a layer, which would reduce the need for mere parameter increment. The conventional widening of networks by having more filters in each layer introduces a quadratic increment of parameters. Having multiple parallel convolutional/dense operations in each layer solves this problem, but without any context-dependent allocation of input among these operations: the parallel computations tend to learn similar features making the widening process less effective. Therefore, we propose the use of multi-path neural networks with data-dependent resource allocation from parallel computations within layers, which also lets an input be routed end-to-end through these parallel paths. To do this, we first introduce a \emph{cross-prediction} based algorithm between parallel tensors of subsequent layers. Second, we further reduce the routing overhead by introducing feature-dependent \emph{cross-connections} between parallel tensors of successive layers. Using image recognition tasks, we show that our multi-path networks show superior performance to existing widening and adaptive feature extraction, even ensembles, and deeper networks at similar complexity.} 
 
\keywords{Multi-path networks, Data-dependent routing, Dynamic routing, Image recognition }

\maketitle

\section{Introduction}
\label{se:intro}

\begin{figure}[t]
	\begin{center}
	\begin{subfigure}{0.32\columnwidth}
		\includegraphics[width=\linewidth, height=1.2in]{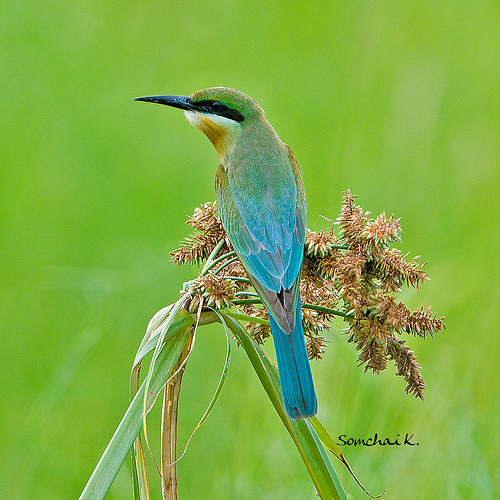}
		\caption{}
		\label{fig:humming1}
	\end{subfigure}
	\begin{subfigure}{0.32\columnwidth}
		\includegraphics[width=\linewidth, height=1.2in]{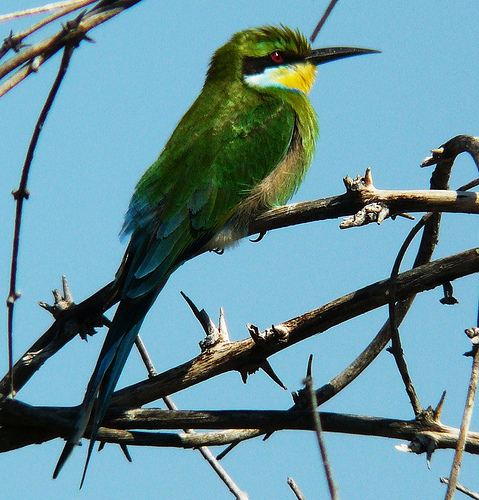}
		\caption{}
		\label{fig:humming2}
	\end{subfigure}
	\begin{subfigure}{0.32\columnwidth}
	    \includegraphics[width=\linewidth, height=1.2in]{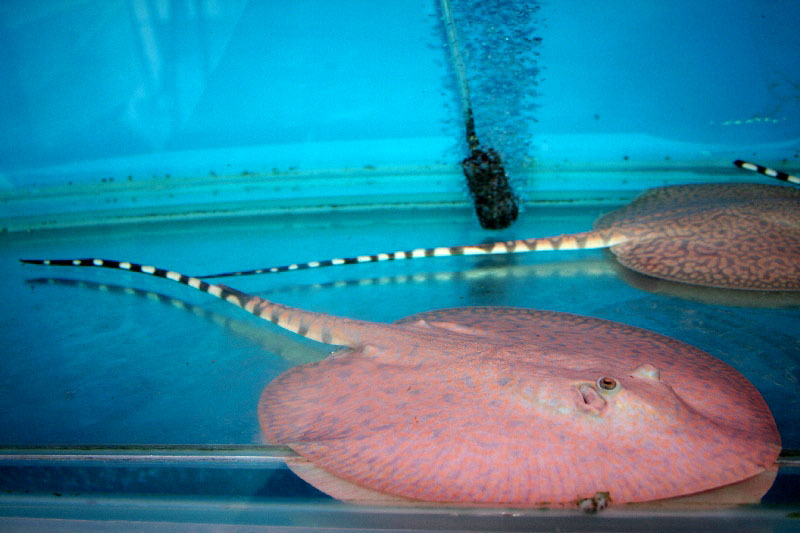}
	    \caption{}
	    \label{fig:madu}
	\end{subfigure}
	\end{center}
	\caption{Intuition for sharing resources at multiple levels of layers: Three samples from the ILSVRC 2012~\citep{ILSVRC15} validation set. The First two images show two hummingbirds, and the third image shows an electric ray. Images b and c share similar low-level features, such as the dominant color, whereas images a and b share similar abstract information, such as body pose. Therefore, processing images b and c feature maps together in the initial layers of a multi-path network and processing images a and b feature maps together in deeper layers might yield better overall performance.}
	\label{fig:intro}
\end{figure}

It is a common practice to increase the depth of a neural network to gain more performance in a given task \citep{resnet, preact-resnet, fitnet, vgg, inception}. While the effective utilization of increasing the depth of a network is well established \citep{resnet, preact-resnet, fitnet}, the efficient utilization of resources in a given layer, i.e., rich feature extraction within a layer, has not been explored well. Having many layers along the depth of a network that are separated by non-linear activations enables a network to approximate very complex distributions. While this is very important to learn in a complex dataset, it is also intuitive to have rich feature extraction processes in each layer, which would improve the network's utility. Furthermore, having a lesser depth reduces the difficulty in learning the network parameters.

The naive way to have richer layer-wise feature extraction is to increase the number of convolutional filters or dense nodes in each layer \citep{wideresnet}. This leads to a quadratic increase in the total number of parameters in terms of the width of the network, which is inefficient. As opposed to this conventional widening, it is efficient to use multiple parallel operations in a given layer \citep{inception, resnetxt} which limits the increment of parameters to be linear in terms of the width. Another approach, model ensembling \citep{alexnet,vgg}, feeds the same image to multiple independent neural networks and amalgamates each network's response. Another class of work feeds different versions of the same image created by different pre-processing mechanisms to multiple independent networks \citep{ciregan2012multi, wang2015multi}. However, without a context-dependent resource allocation from the parallel computations, these widening processes tend to learn redundant information raising questions on the overall effectiveness of having parallel operations. In summary, the existing widening is not fully effective in terms of parallel resource utilization.

To this end, we are inspired by the philosophy of solving a complex problem by breaking the input space into sub-spaces and fitting a group of simpler surfaces in those sub-spaces instead of fitting a single complex surface \cite{friedman1991multivariate, breiman2017classification}. This philosophy can be adopted to network learning by partitioning the input space into sub-spaces, employing specialist experts in each sub-space to extract features, and weighing each expert's output to derive the final prediction \cite{jacobs1991adaptive, jordan1994hierarchical}. Such a divide-and-conquer approach should ideally consist of input-dependent gating/routing mechanisms to delegate computations to sub-experts and accumulate the results. Furthermore, implementing such divide-and-conquer extractors layer-wise would enhance the efficiency of deep feature extraction \cite{eigen2013learning, shazeer2017outrageously}.

Accordingly, consider a particular layer in a multi-path network that contains parallel sets of feature maps (paths). The architecturally parallel families of filters in each path learn independently. Suppose homogeneous feature maps---those that concentrate on similar image features---are already grouped into parallel paths. In that case, each family of filters operating on each path can specialize the feature extraction to the particular context. Such a wise use of parallel resources increases the efficiency of feature extraction with respect to the number of filters used, as multiple small families of dedicated filters may extract a richer combination of features than a large, single set of filters---or even multiple sequential filter sets (deepening). To do such grouping and to allocate incoming parallel tensors to these groups, we need a mechanism that routes between subsequent layers, i.e., to connect the parallel sets of feature maps (tensors) of a particular layer to the parallel paths in the next layer needing cross-connections. This mechanism should further gate these connections according to the context of the previous layer's tensors, so they get routed to the next layer adaptively. The parallel paths would be able to allocate resources efficiently with carefully designed routing.

It is also important to have such routing mechanisms throughout the depth of the multi-path network \cite{eigen2013learning, shazeer2017outrageously}, preferably for each segment of layers, rather than allocating inputs to parallel paths at the very first layer and concatenating the outputs of the final layer. This is because the \emph{context} of an image is captured throughout the depth of the neural network, with each depth segment focusing on a different level of abstraction of the image \cite{alexnet, albawi2017understanding, erhan2009visualizing}. Therefore, in each layer, the homogeneous grouping of feature maps could be different from each other. We interpret an image's context as a cumulative detail that is not limited to the class. An image's context, at the lowest level, might represent the overall color, structure of edges, etc., whereas, at deeper levels, more abstract information, such as body pose or even the class. In addition, the real image context, which matters in the given task, might differ from the human interpretation \citep{kahatapitiya2019context}. Therefore, routing captures context at different levels of abstraction distributed along the depth of the network.

Thus, when a multi-path network with context-wise path allocation learns a particular task, images that get similar resource allocation in a particular depth might get a different allocation at another depth. For example, consider the three images from ILSVRC2012 \citep{deng2009imagenet} dataset shown in Figure \ref{fig:intro}. Image \ref{fig:humming1} shows a hummingbird sitting on a green bench where the background is grass. Image \ref{fig:humming2} is also a hummingbird but sitting on a thorny bench with the background as the sky. Image \ref{fig:madu} is an electric ray in the water. If we consider a shallow layer (low-level) detail such as the overall color of the image, image \ref{fig:humming2} and \ref{fig:madu} are similar to each other,  whereas image \ref{fig:humming1} is different. However, if we consider an abstract detail such as the body pattern, image \ref{fig:humming1} and \ref{fig:humming2} are similar, and image \ref{fig:madu} is different. Therefore, in the initial layers of a multi-path network, image \ref{fig:humming2} and image \ref{fig:madu} might get similar path allocations, and in deeper layers, image \ref{fig:humming1} and \ref{fig:humming2} might get similar path allocations. To accommodate such different groupings according to the nature of features in each layer, we need routing mechanisms throughout the depth of a multi-path network.

In this paper, we introduce novel layer-wise routing mechanisms to softly route an input image among the parallel paths in each layer of a multi-path network. The proposed data/feature-dependent routing algorithms delegate the data flow from the parallel sets of feature maps in a given layer to the parallel paths in the next layer. Such routing layers facilitate all possible connections between two subsequent layers of parallel tensors and adaptively weight those connections with feature-dependent gates. The main contributions of this paper are two-fold as follows, 
\begin{itemize}
\item We first propose a cross-prediction-based algorithm. Each tensor in a given layer of parallel tensors predicts all the following layer tensors and its routing probabilities (gates) to each following layer tensor. Each of the next layer's parallel tensors is constructed by summing the predictions made by previous layer tensors to it weighted by the corresponding gates. 
\item We further propose a cross-connection-based algorithm, where each tensor in a given layer of parallel tensors only computes its routing probabilities (gates) to each following layer tensor. Each following layer tensor is constructed by directly summing the previous layer tensors weighted by the corresponding gates. This design reduces the routing overhead drastically while maintaining performance. 
\end{itemize}
We show that the proposed multi-path networks exhibit superior performance to existing deepening, widening, and adaptive feature extraction methods. Further, we empirically justify the nature of context-dependent resource allocation and gate activation. This paper extends the work carried out by Tissera \etal 2019 \cite{tissera2019context} and Tissera \etal 2020 \cite{tissera2021feature}.

\section{Related Work}
\label{se:related_work}
Convolutional neural networks with many layers along the depth have proven excellent performance in the supervised learning domain \cite{alexnet, vgg, inception}, surpassing conventional shallow neural networks \cite{lenet, rumelhart1986learning}. However, having too many layers in a conventional neural network leads to performance degradation \cite{resnet}. Residual Networks (ResNets) \cite{resnet} mitigate this issue by using residual blocks, which allow the gradients to flow to the initial layers with less attenuation through residual pathways. Identity mappings in residual networks \cite{preact-resnet} further clear the residual pathways enabling the training of very deep networks without gradient attenuation possible. However, these deepening approaches mainly focus on clearing the gradient flow pathways to efficiently train very deep networks, while it is also intuitive to improve the feature extraction process layer-wise.

The conventional width enhancement of convolutional neural networks by increasing the number of filters \cite{wideresnet} or fully-connected nodes in each layer is inefficient as the added complexity outweighs the performance gain. Also, width increment results in quadratic parameter increment, which is inefficient. In contrast, ResNeXt \cite{resnetxt} and Inception networks \cite{inception-v4, inception, szegedy2016rethinking} use parallel operations in a given layer which limits the parameter increment to a linear scale. However, there is no context-dependent allocation of input feature maps among these parallel computations; hence,  parallel paths tend to learn similar information. Model ensembling \cite{alexnet, vgg}, where multiple networks compute independent responses of the same input to compute the final output, is also subjected to this feature redundancy.

Instead of feeding the same input to multiple networks, it is more intuitive to feed different versions of the same input to parallel networks. Ciregan \etal (2012) \cite{ciregan2012multi} showed that having multi-column networks, where each set of columns is fed with inputs pre-processed in different ways, leads to improved performance. Wang (2015) \cite{wang2015multi} proposed a similar approach of using multi-path networks with different versions of input fed to different paths. However, these approaches do not connect parallel paths along the depth; instead, these parallel columns learn in isolation. Since each path only focus on learning from a different version of the same input, there is no context-dependent allocation of parallel resources. To have a multi-path network do such allocation layer-wise, we need connections between parallel computations throughout the depth of the network. 

Cross-Stitch Networks \cite{cross_stich} use weighted cross-connections between parallel networks, where the weighing coefficients of the cross-connections are learned independently and are static during inference. Such work aims to determine the fixed mix of task-specific and shared resources in a parallel-path network to perform multiple tasks for a single input (e.g., semantic segmentation and surface normal estimation), referred to as multi-task learning \cite{caruana1997multitask, thung2018brief, crawshaw2020multi}. Sluice networks \cite{sluice} further add weighted shortcuts along the depth of each network in addition to the layer-wise cross-connections. NDDR-CNN \cite{nddr-cnn} further generalizes the motives of both Cross-Stitch Networks and Sluice networks by using 1$\times$1 convolutions for cross-computations and skip-connections on resized feature maps at different depths (NDDR-CNN Shortcut Network). These multi-task learning networks perform distinct tasks on the same input, where they specifically focus on sharing information learning between the distinct tasks, each learned by a specific network. Hence, the weights governing the resource sharing between parallel networks can be learned independently (static during inference). In contrast, we focus on delegating information learning to parallel paths/operations to learn one task, where it is vital to dynamically compute resource allocating weights based on the input. Therefore, in our case, the weights of such cross-connections should depend on the input features.  

Our work is closely related to existing adaptive feature extraction methods. We use the term adaptive feature extraction because, in those methods, the primary feature extraction process is supported by additional parametric or non-parametric functions. These functions are computed on the inputs to the network \cite{ha2016hypernetworks, cai2021dynamic} or the inputs to each layer \cite{emrouting, hu2018gather, hu2017squeeze, sabour2017dynamic, wang2019eca, convnet-aig, blockdrop, srivastava2015highway, rao2018runtime, wang2018skipnet, chen2021multipath, zhang2022resnest, yu2021path}. Such adaptive functions allow those networks to be flexible to the input context, making the network more dynamic during inference. 
Hypernetworks \cite{ha2016hypernetworks} include a smaller network embedded inside the main network to predict the weights of the main network. Squeeze-and-excitation networks (SENets) \cite{hu2017squeeze} introduce a learnable re-calibration of each convolutional channel, commonly known as channel-wise attention. This channel-wise attention has been subsequently adopted to improve existing networks by channel re-calibration, e.g., MFR-DenseNet \cite{chen2021multipath} improving DenseNets \cite{huang2017densely}, and ResNeSt \cite{zhang2022resnest} improving ResNeXt \cite{resnetxt}. Highway Networks \cite{srivastava2015highway, highway2} propose using gates to learn to regulate the flow of information along the depth of the network to effectively train deep models. ConvNet-AIG \cite{convnet-aig}, BlockDrop \cite{blockdrop}, and SkipNet \cite{wang2018skipnet} introduce data-dependent selection criteria of residual blocks in a ResNet \cite{resnet} according to the nature of the input. However, these approaches mainly utilize a common path for the main flow of information end-to-end, although the weights might vary. In contrast, our model has parallel paths with different weights in each path, enabling the model to vary the main flow of information through a selected combination of parallel resources in each layer according to the context. Ours facilitates context-dependent soft selection and sharing of resources. 

Mixture of experts \cite{jacobs1991adaptive, jordan1994hierarchical} partition the input space to sub-spaces and data-dependently selects specialist experts extract features in each sub-space. Although initial work only used entire models as experts, subsequent work introduced layer-wise mixtures of experts \cite{eigen2013learning, shazeer2017outrageously}. In particular, sparsely-gated mixture of experts \cite{fedus2022review, chen2022towards} have achieved a significant advancement recently in domains such as natural language processing \cite{shazeer2017outrageously, lepikhin2020gshard, fedus2021switch} and vision \cite{riquelme2021scaling, wu2022residual}. However, these sparse mixtures of experts involve hard allocation of inputs to selected experts, hence, often need large amount of data and heavily depend on network engineering across parallel devices during the training phase. In contrast we use soft allocation of parallel resources which supports single device backpropagation. 

\section{Cross-Prediction-based Routing}
\label{se:cp}

To build end-to-end routing in a parallel-path network, we should build a layer-wise routing mechanism to route between subsequent layers carrying parallel tensors in each. I.e., given a layer of parallel tensors, we need a mechanism to construct the next layer of parallel tensors. This mechanism should allow gated coupling between tensors in the two layers so that any tensor in the first layer can be routed to any tensor in the next layer. In our cross-prediction-based algorithm, each tensor among parallel tensors in a particular layer performs convolutional or dense predictions for each of the tensors in the next layer. In addition, each tensor in the former layer also predicts the probabilities (denoted by gates) of that particular tensor being routed to each of the next layer tensors. Each of the next layer parallel tensors is constructed by adding together the predictions made to it, which are weighted by corresponding gates.

Suppose the inputs to a routing layer consist of $m$ tensors [$\mathbf{X}_{i=1,...,m}$], and the routing layer outputs $n$ tensors [$\mathbf{Y}_{j=1,...,n}$]. First, each tensor in inputs performs predictions for each tensor in the outputs. The prediction $\mathbf{U}_{ij}$, which is made by tensor $i$ in inputs ($\mathbf{X}_i$) to tensor $j$ in outputs ($\mathbf{Y}_j$), is a linear, learnable transformation, which can be denoted as,
\begin{equation}
\nonumber
	    \mathbf{U}_{ij} = \mathrm{W}_{ij}\mathbf{X}_i + b_{ij},
	    \label{eq:u_ij}
\end{equation} 
where $\mathrm{W}_{ij}$ and $b_{ij}$ correspond to weight and bias terms, respectively. If $\mathbf{X}_i$ is a 3-dimensional tensor ($\mathbf{X}_i \in \mathbb{R}^{W\times H\times C}$), this corresponds to a convolution. 

In addition, each $\mathbf{X}_i$ predicts an $n$-dimensional vector of gate values $\mathbf{G}_i$, which represents the $n$ probabilities of $\mathbf{X}_i$ being routed to each $\mathbf{Y}_j$, i.e., $\mathbf{G}_i$ can be expressed as $[g_{i1}, \dots, g_{in}]$, where, $g_{ij}$ corresponds to the scalar gate value connecting $\mathbf{X}_i$ to $\mathbf{Y}_j$. $\mathbf{G}_i$ can be calculated by a non-linear parametric computation on $\mathbf{X}_i$, preferably two dense operations separated by $ReLU$ activation. However, If $\mathbf{X}_i$ is 3-dimensional, this occupies a significant amount of parameters. Therefore, if $\mathbf{X}_i$ is 3-dimensional, to calculate $\mathbf{G}_i$, we first feed $\mathbf{X}_i$ to a global average pooling operation, to produce $1\times 1\times C$ latent channel descriptor $\mathbf{Z}_i$ \citep{hu2017squeeze, convnet-aig}. Since each channel in a set of convolutional feature maps represents a particular feature of the input, which is searched by a specific filter, global average pooling results in a compressed descriptor that still carries the information about the presence of each feature. Global average pooling regularizes the gating computation by preventing it from overfitting to its input tensor. The $c^{th}$ channel value $(z_i)_c$ of the channel descriptor $\mathbf{Z}_i$ can be obtained as,
\begin{equation}
	{(z_i)}_c = \frac{1}{H\times W}\sum_{a=1}^{H}\sum_{b=1}^{W}{(x_i)}_{a,b,c}.
	\label{eq:gap}
\end{equation}
$\mathbf{Z}_i$ is then fed to a non-linear computation with two fully-connected layers (weights $\mathrm{W}_1$ and $\mathrm{W}_2$), separated by $\mathrm{ReLU}$ activation \citep{glorot2011deep}. This operation yields $n$ latent relevance scores $\mathbf{A}_i$  ($[a_{i1},\dots ,a_{in}]$) representing the relevance of the incoming tensor to the next layer tensors:
\begin{equation}
	\mathbf{A}_i = \mathrm{W}_2(\mathrm{ReLU}(\mathrm{W}_1\mathbf{Z}_i)).
	\label{eq:relevance}
\end{equation}
Finally, we impose $\mathrm{softmax}$ activation on top of the $n$ relevance scores $\mathbf{A}_i$ to calculate gate probabilities $\mathbf{G}_i$:
\begin{equation} 
    \mathbf{G}_i = \mathrm{softmax}(\mathbf{A}_i), \hspace{0.1in} \text{i.e.,}  \hspace{0.1in} g_{ij} = \frac{\mathrm{e}^{a_{ij}}}{\sum_{k=1}^{n} \mathrm{e}^{a_{ik}}} .
    \label{eq:softmax}
\end{equation}

\begin{figure*}[t]
\begin{center}
    \begin{subfigure}[]{0.49\linewidth}
		\includegraphics[width=\linewidth]{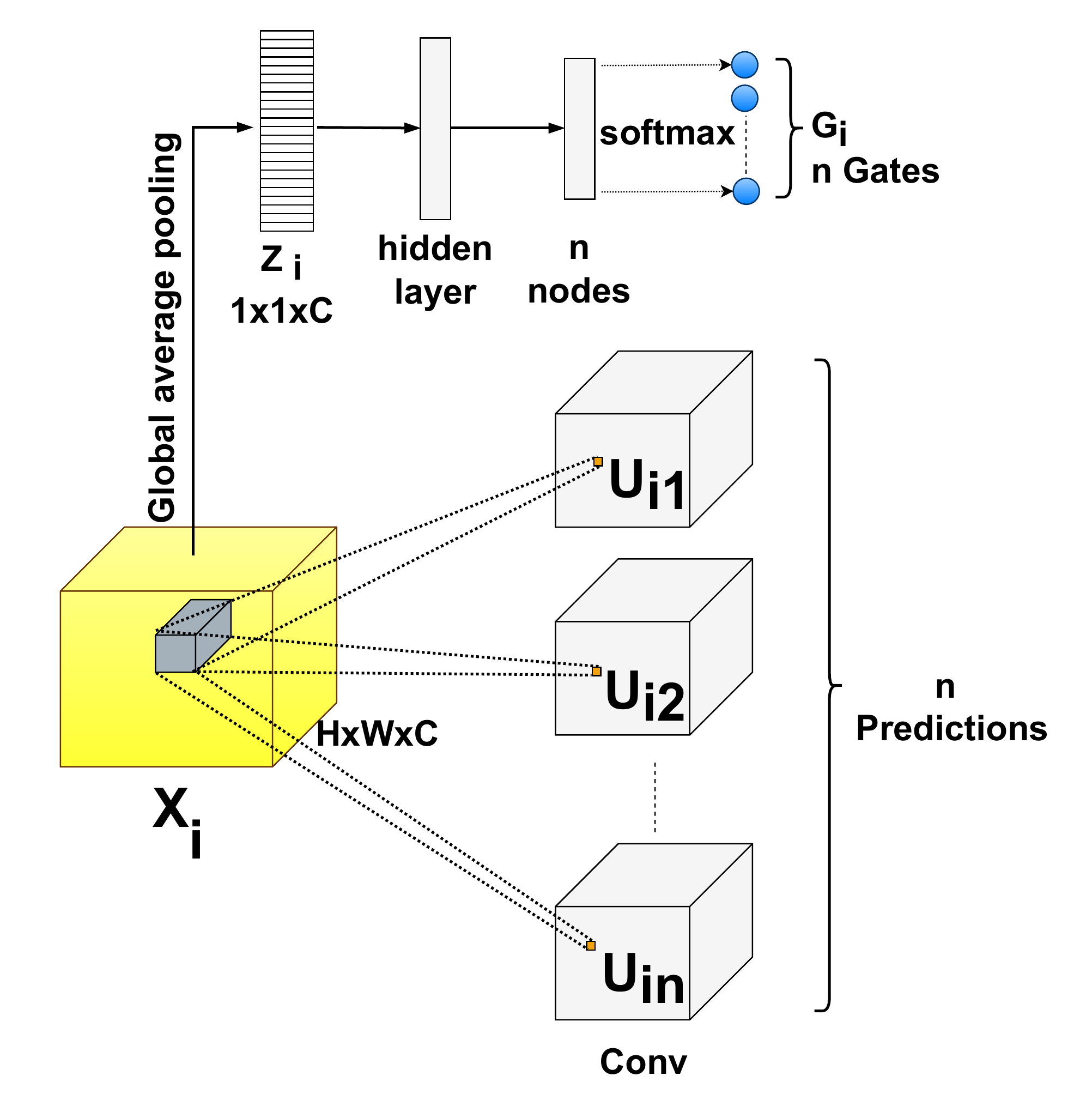} 
		\caption{}
		\label{fig:prediction}
	\end{subfigure}
    \begin{subfigure}[]{0.49\linewidth}
		\includegraphics[width=\linewidth]{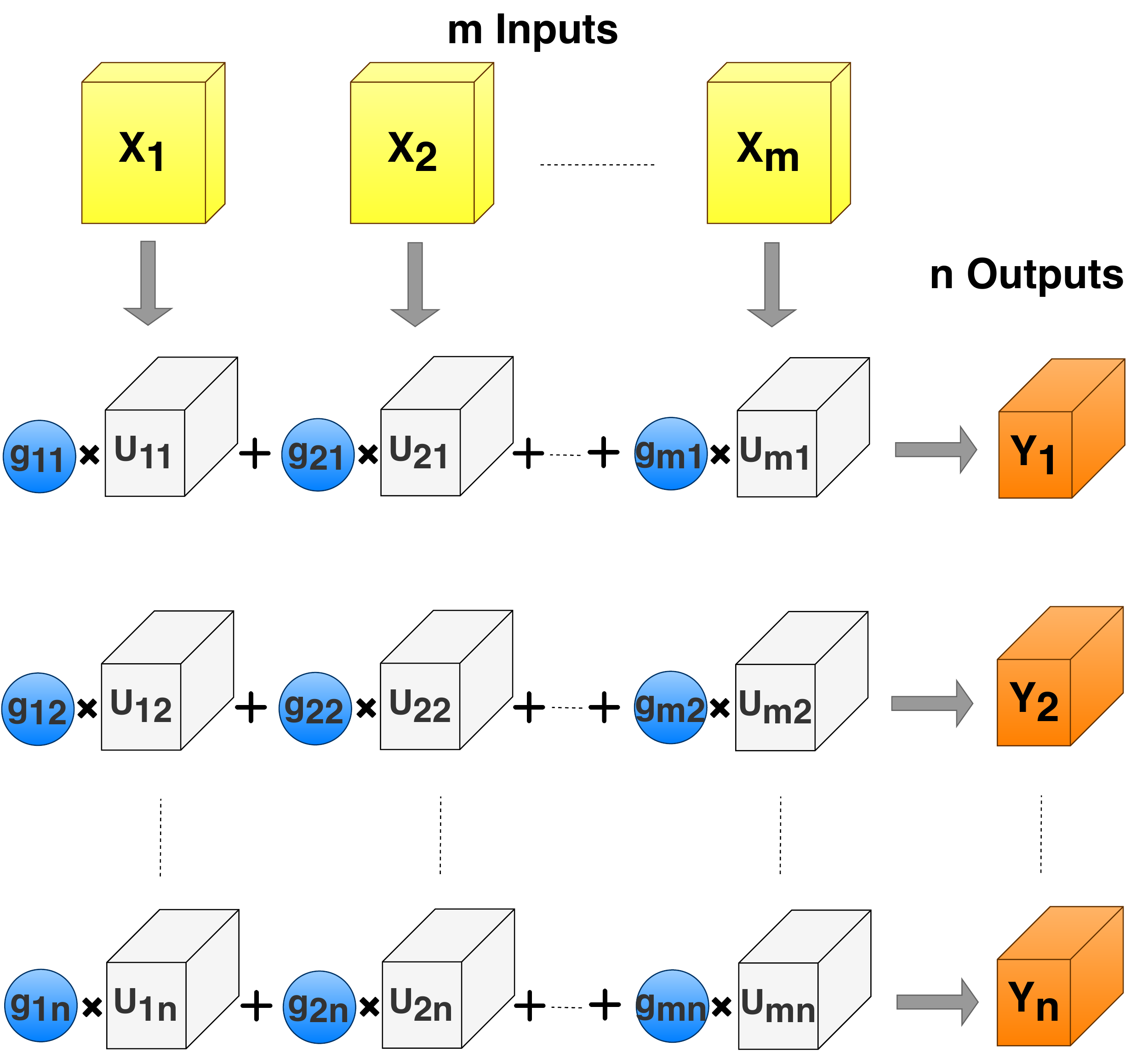} 
		\caption{}
		\label{fig:construction}
	\end{subfigure}
\end{center}
\caption{\ref{fig:prediction}: 3-dimensional tensor $\mathbf{X_i}$ in inputs predicting $n$ outputs of routing layer and associated gates. \ref{fig:construction}: Constructing outputs of routing layer based on the predictions and gates computed by all such inputs $\mathbf{X}_{i=1,...,m}$. See Eq. \ref{eq:y_gu}.}
\label{fig:cross-predictions}
\end{figure*}

The activation $\mathrm{softmax}(.)$ returns $n$ scores, which represent the probabilities of $\mathbf{X}_i$ being routed to each output $\mathbf{Y}_{j=1,...,n}$. Figure~\ref{fig:prediction} shows the operations carried out by a 3-dimensional tensor at the input of a routing layer in the prediction phase.
 
Once the cross-predictions $\mathbf{U}_{ij}$ and the gates $\mathbf{G}_i$ are calculated, the outputs of the routing layer are calculated. To construct $j^{th}$ output $\mathbf{Y}_j$, predictions made for $\mathbf{Y}_j$ ($\mathbf{U}_{ij}$, $i=1,\dots, m$) are weighted by corresponding gate values ($g_{ij}$, $i=1,\dots, m$) and added together. We further impose $\mathrm{ReLU}$ activation to the constructed tensor.
\begin{equation}
	\mathbf{Y}_j = \mathrm{ReLU}\left(\sum_{i=1}^{m} (g_{ij}\times \mathbf{U}_{ij})  \right) .
	\label{eq:y_gu}
\end{equation}
This adaptive re-calibration of the predictions made by input tensors to construct the output tensors shares a similar intuition of attention introduced in SENets \citep{hu2017squeeze}. We intend to use such an attention mechanism to softly route information along different paths.

\begin{figure*}[t]
\begin{center}
	\includegraphics[width=0.98\linewidth]{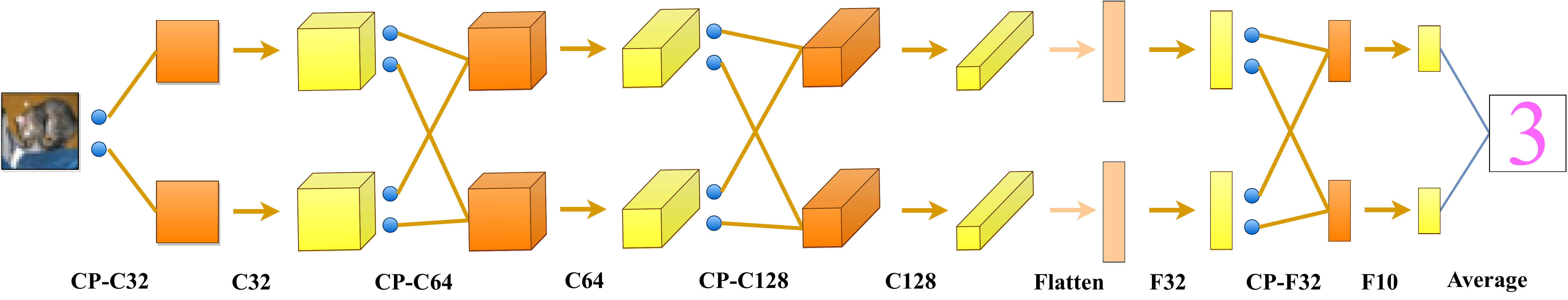} 
\end{center}
\caption{Two-path CNN for image classification with cross-prediction-based routing (referred to as BaseCNN-2-CP in the paper). CP-C$n$ denotes a cross-prediction-based routing layer where the cross predictions are convolutions with $n$ filters. Similarly, CP-F$n$ denotes a routing layer with dense cross-predictions of $n$ nodes. C$n$ denotes a forward layer where parallel computations are convolutions, each with $n$ filters. F$n$ denotes a forward layer with parallel dense layers, each containing $n$ output nodes.}
\label{fig:mpn_cp}
\end{figure*}

Figure \ref{fig:mpn_cp} shows a two-path convolutional neural network with our routing added at selected locations, which is referred to as BaseCNN-2-CP later. The routing process between two layers with $m$ inputs and $n$ outputs is illustrated in Figure \ref{fig:cross-predictions}. There, Figure \ref{fig:prediction} shows a particular tensor among the inputs to a routing layer predicting next layer tensors and its coupling probabilities to them. Figure \ref{fig:construction} shows the construction of the outputs of the routing layer from the predictions and gates calculated by previous layer tensors. Algorithm \ref{alg:cross-predictions} further explains the routing between two layers. 

\begin{algorithm}[t]
	\caption{Cross-Prediction-based routing between inputs and outputs of a routing layer.}
	\label{alg:cross-predictions}
	\begin{algorithmic}
		\State {\bfseries Input: $\mathbf{X}: $ [$\mathbf{X_i}$ for $i=1,2,\dots,m$]}
		
		\State {\bfseries Predictions from current layer:}
		\For{$i=1$ {\bfseries to} $m$}
		\For{$j=1$ {\bfseries to} $n$}
		\State $\mathbf{U}_{ij} \leftarrow \mathrm{W}_{ij}\mathbf{X}_i + b_{ij}$ 
		\EndFor
		
		\State {\bfseries Gate Computation on $\mathbf{X_i}$:}
		\State $\mathbf{Z}_i \leftarrow \mathrm{global\_average\_pooling}(\mathbf{X}_i)$
		\State $\mathbf{A}_i = [a_{i1},\dots ,a_{in}]  \leftarrow 
		\mathbf{W}^i_2(\mathrm{ReLU}(\mathbf{W}^i_1\boldsymbol{Z}_{i}))$
		\State $\mathbf{G}_i = [g_{i1},\dots ,g_{in}] \leftarrow \mathrm{softmax}(\mathbf{A}_i)$ 
		\EndFor	
		
		\State {\bfseries Construction of outputs:}
		\For{$j=1$ {\bfseries to} $n$}
		\State $\mathbf{Y}_j \leftarrow \mathrm{ReLU}(\sum_{i=1}^{m} (g_{ij}\times \mathbf{U}_{ij}))$
		\EndFor
		
		\State {\bfseries Output: $\mathbf{Y}: $ [$\mathbf{Y}_j$ for $j=1,2,\dots,n$]} 
	\end{algorithmic}
\end{algorithm}

We insert these routing layers between selected layers in multipath networks (Figure~\ref{fig:mpn_cp}), enabling other layers to have independent parallel paths to learn in an isolated manner. Adding one routing layer increases the effective depth of the network by one layer due to the cross-predictions being convolutional or dense operations. Since the output layer tensors are combinations of linear operations, it is important to impose a non-linear $ReLU$ activation before feeding the parallel tensors to the next feed-forward computation. In the final layer, the parallel feature maps are averaged to produce a single output.

However, since each tensor in a given layer predicts each tensor in the subsequent layer in terms of a convolution or a dense operation (cross-predictions), the number of parameters employed in the routing process between two layers quadratically rises with the number of parallel paths. Having such an amount of routing overhead is not efficient. Therefore, to limit the routing overhead increment to be linear with the number of parallel paths, we introduce cross-connection-based routing.

\section{Cross-Connection-based Routing}
\label{se:cc}

Cross-connection-based routing is similar to the above-explained cross-prediction-based routing (Sec.~\ref{se:cp}). Instead of weighing cross-predictions that involve either dense or convolutional operations, it weights the input tensors of the routing layer to construct output tensors. This way, the quadratic increment of routing overhead with the number of parallel paths is overcome. The routing overhead now only contains the small number of parameters added from the non-linear gate computations. Also, a routing layer now becomes a mere cross-connecting layer and does not carry weights which are contributed to learning the main task. Therefore, inserting cross-connections between layers in a multi-path network facilitates soft routing without the disadvantage of increasing the effective depth.

\begin{figure*}[t]
\begin{center}
	\includegraphics[width=0.98\linewidth]{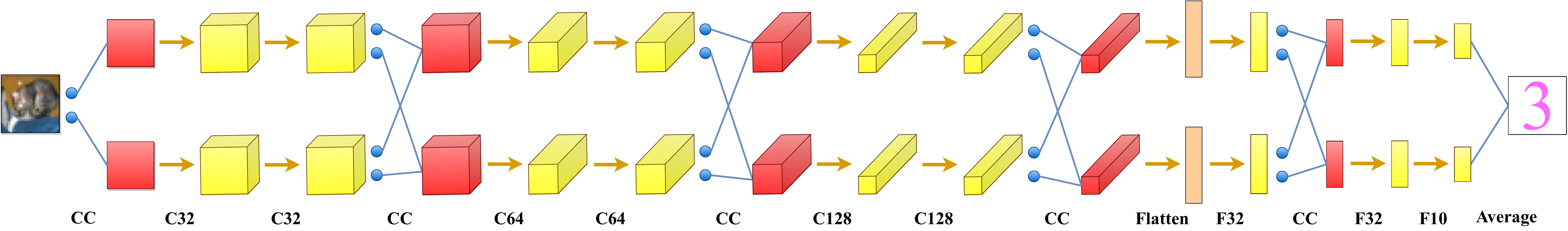} 
\end{center}
\caption{Two-path CNN for image classification with adaptive cross-connections inserted at selected locations (referred to as BaseCNN-2-CC in the paper). CC denotes a cross-connecting layer where the gates and connections are shown by blue circles and edges, and the outputs of cross-connecting layers are shown in red boxes. C$n$ and F$n$ denote forward convolutional and dense layers, respectively, as in Fig. \ref{fig:mpn_cp}. The outputs of such forward layers are depicted by yellow boxes. Since the cross-connections are mere weighted connections, adding cross-connecting layers does not increase the effective depth of the network.}
\label{fig:mpn_cc}
\end{figure*}

\begin{figure*}[t]
\begin{center}
   \includegraphics[width=0.8\linewidth]{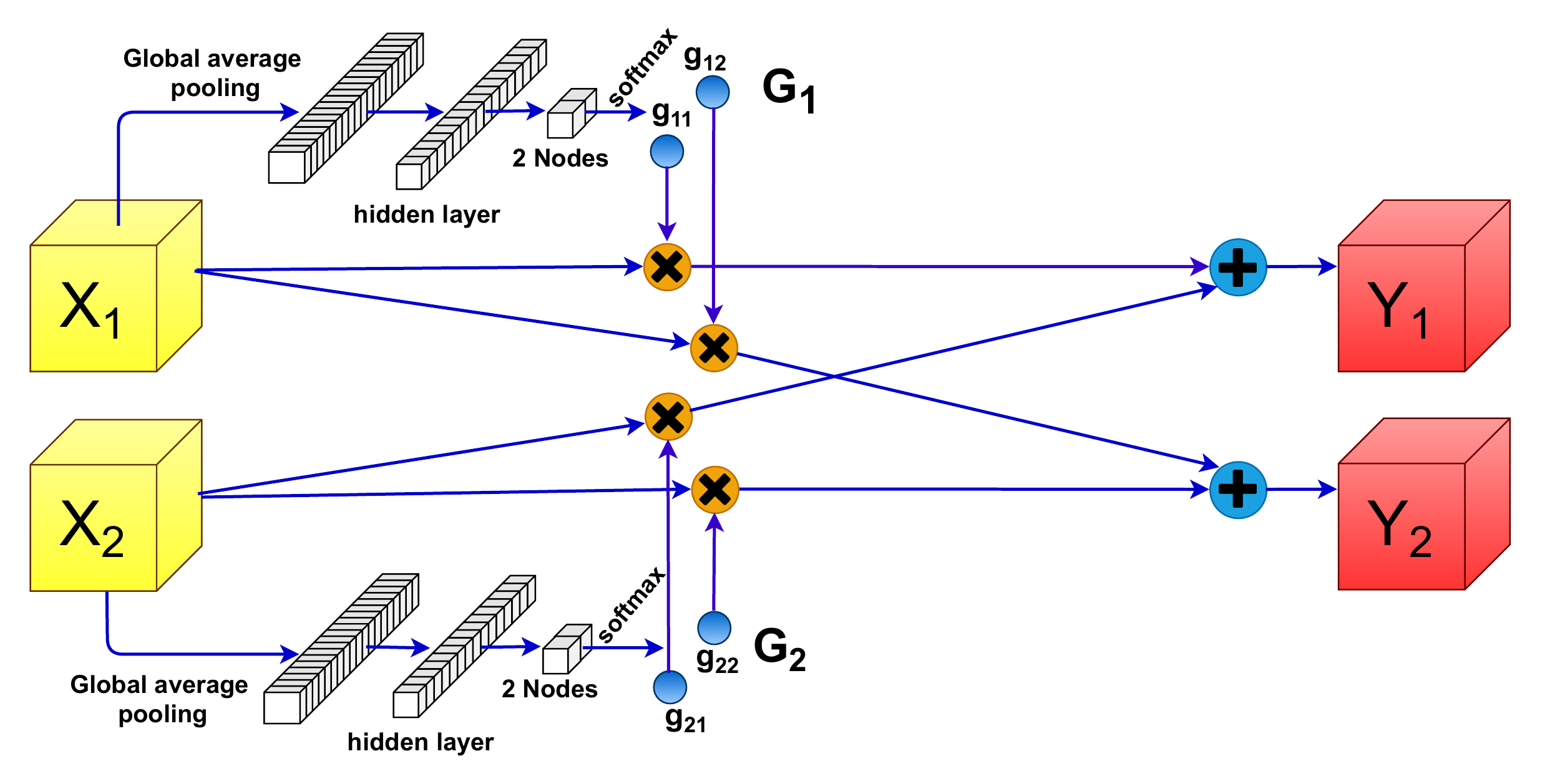}
\end{center}
\caption{The Cross-connecting process between two layers, each with two parallel tensors. The gates which weigh the connections are computed from the input tensors by learnable parametric computations.}
\label{fig:cross-connections}
\end{figure*}

Given the $m$ inputs [$\mathbf{X}_{i=1,\dots,m}$], to produce the $n$ outputs [$\mathbf{Y}_{j=1,...,n}$], each $\mathbf{X}_i$ computes the gate vector $\mathbf{G}_i$ ($[g_{i1}, \dots, g_{in}]$) as depicted by Eq.~\ref{eq:gap}, Eq.~\ref{eq:relevance} and Eq. \ref{eq:softmax}. Given the gates, the algorithm next computes each $\mathbf{Y}_j$ output by summing the inputs [$\mathbf{X}_{i=1,\dots,m}$] each weighted by the corresponding gate $g_{ij, i=1,\dots,m}$:
\begin{equation} 
	    \mathbf{Y}_j  =  \sum_{i=1}^{m} (g_{ij}\times \mathbf{X}_i).
	    \label{eq:y_gx}
\end{equation} 
Since we directly connect inputs to construct outputs, the output tensor dimensions are the same as the inputs. Figure~\ref{fig:mpn_cc} shows a two-path CNN with routing layers inserted at selected locations. It is referred to as BaseCNN-2-CC later. Figure~\ref{fig:cross-connections} shows the cross-connecting process between two layers carrying two parallel tensors in each. Algorithm \ref{alg:cross-connections} illustrates the adaptive cross-connecting process.

\begin{algorithm}[t]
	\caption{Cross-connection-based routing between two adjacent layers with $m$ input and $n$ output sets of feature maps, respectively.}
	\label{alg:cross-connections}
	\begin{algorithmic}
		\State {\bfseries Input:} 
		\State $\mathbf{X}$: [$\mathbf{X}_i$ for $i=1,\dots,m$]
		\State {\bfseries Calculating gate values:}
		\For{$i=1$ {\bfseries to} $m$}
		\State $\mathbf{Z}_i \leftarrow \mathrm{global\_average\_pooling}(\mathbf{X}_i)$
		\State $\mathbf{A}_i = [a_{i1},\dots ,a_{in}]  \leftarrow  
		\mathbf{W}^i_2(\mathrm{ReLU}(\mathbf{W}^i_1\boldsymbol{Z}_{i}))$
		\State $\mathbf{G}_i = [g_{i1},\dots ,g_{in}] \leftarrow \mathrm{softmax}(\mathbf{A}_i)$ 
		\EndFor	
		\State {\bfseries Construction of outputs:}
		\For{$j=1$ {\bfseries to} $n$}
		\State $\mathbf{Y}_j \leftarrow  \sum_{i=1}^{m} (g_{ij}\times \mathbf{X}_{i})$
		\EndFor
		\State {\bfseries Output:} 
		\State $\mathbf{Y}$: [$\mathbf{Y}_j$ for $j=1,\dots,n$]
	\end{algorithmic}	
\end{algorithm}

We further illustrate the cross-connecting process by matrix form to show the pixel-wise operations. Consider a set of 3-dimensional input tensors [$\mathbf{X}_{i=1,...,m}$] and output tensors [$\mathbf{Y}_{j=1,...,n}$]. Let's denote the pixel value at the location $(a,b,c)$ of $\mathbf{X}_i$ as $(x_i)_{a,b,c}$, and $\mathbf{Y}_j$ as $(y_j)_{a,b,c,}$. The set of output pixels at $(a,b,c)$ are therefore, 
\begin{equation}
    \begin{bmatrix}
    (y_1)_{a,b,c} \\
    \vdots \\
    (y_n)_{a,b,c}   \\
    \end{bmatrix}  = 
    \begin{bmatrix}
    g_{11} & \cdots &  g_{m1}   \\
    \vdots & \ddots & \vdots    \\
    g_{1n} & \cdots &  g_{mn} \\
    \end{bmatrix}
    \begin{bmatrix}
    (x_1)_{a,b,c} \\
    \vdots \\
    (x_m)_{a,b,c}   \\
    \end{bmatrix}.
    \label{eq:y_gx2}
\end{equation}
This formulation is similar to Cross-Stitch Networks \citep{cross_stich}. However, their coupling coefficients $g_{ij}$ are independently trained weights. Thus, the coupling coefficients only allow learning the mix of shared and task-specific representations to perform multiple tasks on a single input which is fixed during inference. In our algorithm, $g_{ij}$s are produced by a parametric computation on inputs $\mathbf{X}_i$ themselves, using the channel-wise attention mechanism \citep{hu2017squeeze}. Such an adaptive gate computation allows dynamic change in the mix of context-specific and shared representations to perform a given task according to the nature of the diverse input.

\section{Back-propagating Gradients through Cross-Connections}
\label{se:backprop}

We saw in Sec.~\ref{se:cc} that cross-connections facilitate context-specific soft routing. Training a network with cross connections need backpropagation of gradients through them. The backpropagation through a cross-connecting layer, represented by Eq.~\ref{eq:y_gx} and \ref{eq:y_gx2}, is not straightforward as in Cross-Stitch networks \citep{cross_stich} where the coupling coefficient matrix consists of independently learned weights. In this case, the elements in the gating matrix $\mathbf{G}$ are constructed from the input $\mathbf{X}$ itself. Therefore the gradient flow to each input $\mathbf{X}_i$ consists not only of the direct gradient weighted by the gate element but also another component from the gate computation. Also, instead of directly optimizing gates, the weights which produce the gates are getting optimized. 

\begin{figure*}[t]
\begin{center}
   \includegraphics[width=\linewidth]{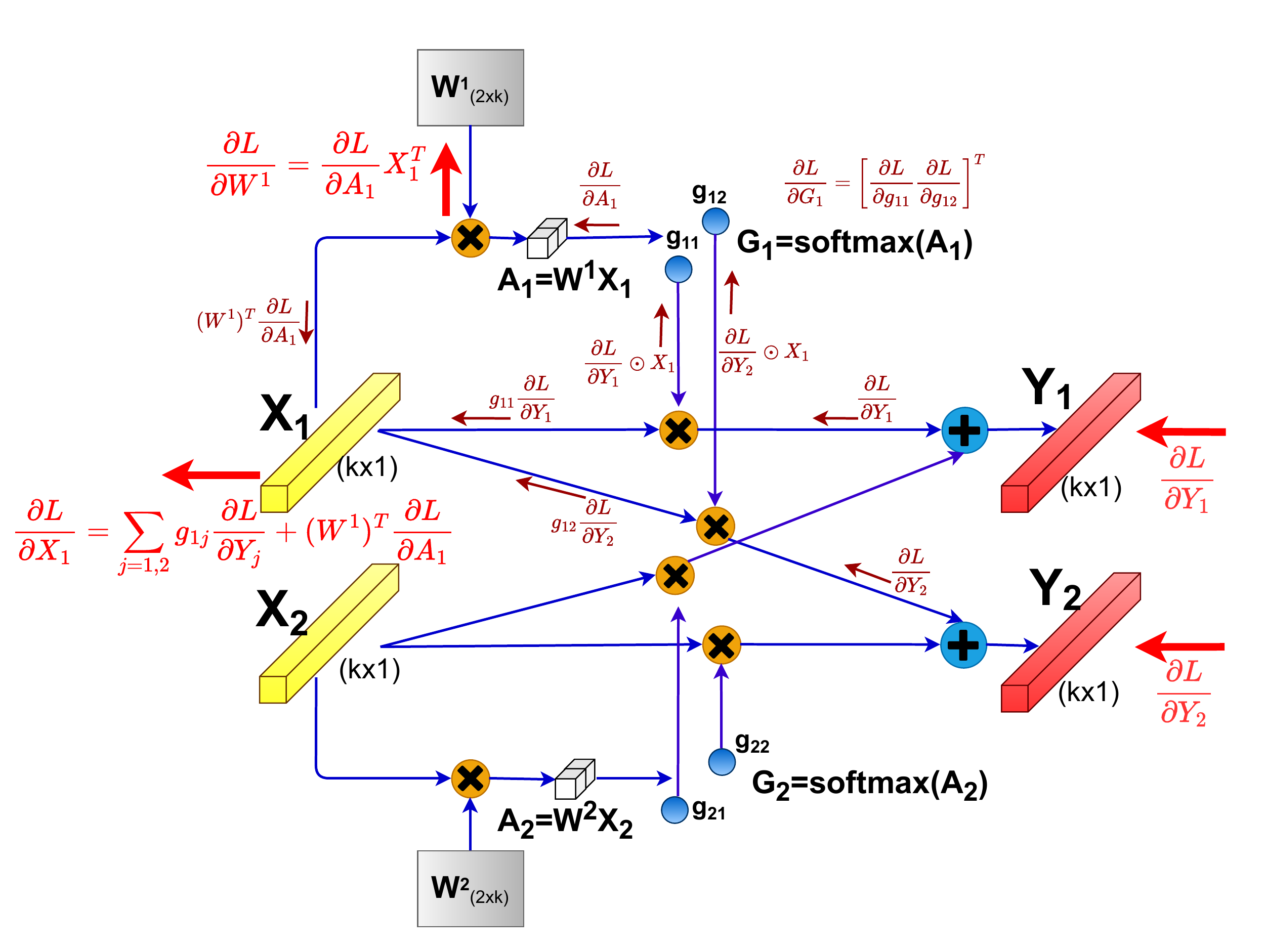}
\end{center}
\caption{The simplified cross-connecting process between two subsequent layers, carrying two parallel tensors in each. Gradient flow to the top tensor in the input layer $X_1$ and its gate computation  weight matrix $W^1$ are shown.}
\label{fig:backprop}
\end{figure*}

For the simplicity of explanation, let's assume that the tensors $\mathbf{X}$ and $\mathbf{Y}$ are $k$-dimensional vectors, and the gate calculation only has a simple fully-connected layer as opposed to Eq.~\ref{eq:gap} and Eq.~\ref{eq:relevance}. Figure \ref{fig:backprop} shows this simplified cross-connecting operation for two-parallel paths. With this simplified operation, calculation of the relevance scores $\mathbf{A}_i$ from each $\mathbf{X}_i$ reduces to, 
\begin{equation}
    \mathbf{A}_i = {\mathbf{W}^i}\mathbf{X}_i , 
	\label{eq:relevance2}
\end{equation}
where $\mathbf{W}^i$ is a $n\times k$ matrix of weights. $\mathbf{G}_i$ is computed by taking $softmax$ of these logits as usual (Eq.~\ref{eq:softmax}). Then, output tensors $\mathbf{Y}_{j\:(j=1\dots n)}$ are constructed as in Eq.~\ref{eq:y_gx}. Here, our goal is to find gradients w.r.t each $\mathbf{X}_{i\:(i=1\dots m)}$ and $\mathbf{W}^i_{(i=1\dots m)}$, given the gradients of loss w.r.t. each output $\mathbf{Y}_{j\:(j=1\dots n)}$. I.e., given $\frac{\partial L}{\partial \mathbf{Y}_{j}}_{j=1\dots n}$, to compute $\frac{\partial L}{\partial \mathbf{W}^i}_{i=1\dots m}$ and $\frac{\partial L}{\partial \mathbf{X}_i}_{i=1\dots m}$. Figure \ref{fig:backprop} shows the flow of gradients to $\mathbf{W}^1$ and $\mathbf{X}_1$ from $\mathbf{Y}_{j(j=1,2)}$ in a two parallel-path cross-connecting operation, which aids understanding the detailed flow of gradients explained below.

It is important to propagate the incoming gradient to each $g_{ij}$ first. The scalar $g_{ij}$ is used to multiply each element of $\mathbf{X}_i$ when producing $\mathbf{Y}_j$ (Eq.~\ref{eq:y_gx}). Therefore, the partial derivative of loss w.r.t. $g_{ij}$ is the summation of the element-wise multiplication between the gradient vector and $\mathbf{X}_i$,
\begin{equation}
    \nonumber
	\frac{\partial L}{\partial g_{ij}} = 
	\sum_{k} \frac{\partial L}{\partial \mathbf{Y}_j} \odot \mathbf{X}_i.
	\label{eq:partial_gij}
\end{equation}
Here, $\odot$ stands for the element-wise multiplication. With all such $\frac{\partial L}{\partial g_{ij}}_{(j=1,\dots,n)}$ derived, we can form $\frac{\partial L}{\partial \mathbf{G}_{i}}$ as an $n$-dimensional column vector,
\begin{equation}
\nonumber
    \frac{\partial L}{\partial \mathbf{G}_{i}} =
    \begin{bmatrix}
    \frac{\partial L}{\partial g_{i1}} &  \cdots  & \frac{\partial L}{\partial g_{in}} \\
    \end{bmatrix}^T.
\end{equation}

Propagating gradients to the relevance scores $\mathbf{A}_i$ involves multiplying the gradients w.r.t $\mathbf{G}_i$ by the partial derivative of gate values w.r.t the relevance scores $\frac{\partial \mathbf{G}_i}{\partial \mathbf{A}_{i}}$, i.e.,
\begin{equation}
\nonumber
    \frac{\partial L}{\partial \mathbf{A}_{i}} = \frac{\partial \mathbf{G}_i}{\partial \mathbf{A}_{i}}^T \frac{\partial L}{\partial \mathbf{G}_{i}} = 
            \left(J^{\mathbf{G}_i}_{\mathbf{A}_i}\right)^T \frac{\partial L}{\partial \mathbf{G}_{i}}.
\end{equation}
Here, $J^{\mathbf{G}_i}_{\mathbf{A}_i}$ is the Jacobian matrix of the softmax derivative,
\begin{equation}
\nonumber
    \frac{\partial \mathbf{G}_i}{\partial \mathbf{A}_{i}} = J^{\mathbf{G}_i}_{\mathbf{A}_i} =
    \begin{bmatrix}
    g_{i1}(1-g_{i1}) &  \cdots  & -g_{i1} g_{in} \\
    \vdots & \ddots & \vdots \\
    -g_{in} g_{i1} &  \cdots  & g_{in} (1 - g_{in}) \\
    \end{bmatrix}.
\end{equation}

The gradients of loss w.r.t. $\mathbf{W}^i$ can now be obtained by propagating the gradient w.r.t $\mathbf{A}_i$ through Eq. \ref{eq:relevance2}. Therefore,
\begin{equation}
    \frac{\partial L}{\partial \mathbf{W}^{i}} = \frac{\partial L}{\partial \mathbf{A}_{i}} \mathbf{X}_i^T
    = \left(J^{\mathbf{G}_i}_{\mathbf{A}_i}\right)^T \frac{\partial L}{\partial \mathbf{G}_{i}} \mathbf{X}_i^T.
\end{equation}
It is also important to calculate the gradient of loss w.r.t $\mathbf{X}_i$ since this is the gradient that is propagated to the previous layer. 
\begin{equation}
    \begin{split}
    \frac{\partial L}{\partial \mathbf{X}_{i}} =  \sum_{j=1}^{n} g_{ij} \frac{\partial L}{\partial \mathbf{Y}_{j}}   +  (\mathbf{W}^i)^T \frac{\partial L}{\partial \mathbf{A}_{i}} \\
    =  \sum_{j=1}^{n} g_{ij} \frac{\partial L}{\partial \mathbf{Y}_{j}}   +  \left(\mathbf{W}^i\right)^T \left(J^{\mathbf{G}_i}_{\mathbf{A}_i}\right)^T \frac{\partial L}{\partial \mathbf{G}_{i}} .
    \end{split}
\end{equation}
Here, the first part of the loss is the direct flow of gradient to $\mathbf{X}_i$ from the multiplication operation between $g_{ij}$ and $\mathbf{X}_i$. The second term reflects the portion of the gradient propagated to $g_{ij}$ from that particular multiplication flowing back to $\mathbf{X}_i$. This residual gradient is due to the attention-like gating mechanism, which produces $g_{ij}$ from $\mathbf{X}_i$ itself.

\section{Image Recognition Performance}
\label{se:experiments}
We conduct various experiments in the image-recognition domain to validate the effectiveness of having parallel paths with data-dependent resource allocation. We first evaluate the impact of having parallel paths in conventional convolutional neural networks. Then, we build custom Residual Networks (ResNets) \citep{resnet} with parallel paths and our routing algorithms. In both cases, we compare our multi-path networks with wide networks, existing adaptive feature-extracting methods, and deeper networks of similar complexity. Among the existing related methods, if the performance of models that carry similar complexity of our multi-path networks are not reported, we build custom models that match our models' complexity.

\subsection{Datasets}
\label{ss:datasets}
We use three image recognition datasets to validate our models and compare them with existing work. CIFAR10 \cite{cifar100} is a 10-class dataset comprising 60k color images of size 32$\times$32. The 60k images are evenly distributed among the ten classes, resulting in 6000 images per class. The training set contains 50k images, and the validation set has 10k images. CIFAR100 \cite{cifar100} is similar to CIFAR10, except for its 60k images are evenly distributed under 100 classes. ILSVRC 2012 Dataset \cite{deng2009imagenet, ILSVRC15} is a large-scale image recognition dataset that contains 1.3M training images and 50k validation images distributed under 1000 categories. Its images are of varying sizes, hence we re-scale them to 256$\times$256.

\subsection{Conventional Convolutional Neural Networks with Parallel Paths}
\label{ss:multi_cnn}

\begin{table*}[t]
\caption{Notations and details of the compared convolutional neural networks: $Cn$ denotes a convolutional layer of $n$ filters. $Fn$ denotes a fully connected layer of $n$ output nodes.}
\label{tab:compared_networks_cnn}
\begin{center}
\begin{tabular}{@{}lr@{}}
\toprule
Network & Structure \\
\midrule
 BaseCNN &  $C32$ $C32$ $C64$ $C64$ $C128$ $C128$ $F32$ $F32$ $F10$ \\
 WideCNN &  $C64$ $C64$ $C128$ $C128$ $C256$ $C256$ $F32$ $F32$ $F10$ \\
 DeepCNN &  $C32$ $C32$ $C64$ $C64$ $C128$ $C128$ $C128$  \\& $C256$  $C256$ $C256$  $F32$ $F32$ $F10$ \\
 BaseCNN-X & BaseCNN--X paths. No routing. \\
 Base Ensemble &  Ensemble of 3 BaseCNNs \\
 All Ensemble &  Ensemble of BaseCNN, WideCNN and DeepCNN \\
 SEBaseCNN &  SENet (\cite{hu2017squeeze}) on BaseCNN \\
 SEDeepCNN &  SENet (\cite{hu2017squeeze}) on DeepCNN \\ 
 Cr-Stitch2 &  Cross-stitch network (\cite{cross_stich}) with 2 parallel BaseCNNs  \\
 NDDR-CNN2 &  NDDR-CNN (\cite{nddr-cnn}) with 2 parallel BaseCNNs  \\
 NDDR-CNN2-shortcut &  NDDR-CNN shortcut net (\cite{nddr-cnn}) with 2 parallel BaseCNNs \\
\midrule
 BaseCNN-X-CP &  BaseCNN--X paths--cross-prediction-based routing \\
 BaseCNN-X-CC&  BaseCNN--X paths--cross-connections \\
\bottomrule
\end{tabular}
\end{center}
\end{table*}

In this section, we add parallel paths to conventional convolutional neural networks and compare them with conventional network widening, deepening and other related networks. Table \ref{tab:compared_networks_cnn} shows the details of the networks we use for this purpose. We choose a 9-layer convolutional neural network (6 convolutional layers and 3 dense layers) as the baseline, denoted as BaseCNN. We build our multi-path networks based on the BaseCNN. 

BaseCNN-X-CP denotes an X-path network with cross-prediction-based routing where each path is similar to a BaseCNN. Figure \ref{fig:mpn_cp} shows BaseCNN-2-CP architecture which uses two parallel paths. Here, $1^{st}$, $3^{rd}$ and $5^{th}$ convolutional layers, and $2^{nd}$ dense layer are replaced by cross-prediction-based routing layers. The first layer is a one-to-many router which connects the input to a given number of tensors. Since cross-predictions are convolutions or dense operations, one routing layer adds one layer to the effective depth of the network. Therefore, to construct the BaseCNN-X-CP network, we replace the selected layers in parallel-path BaseCNN with the routing layers to maintain the same depth as BaseCNN. Finally, the outputs of the last layer of parallel dense operations are averaged to produce the final prediction.

BaseCNN-X-CC is an X-path network with adaptive cross-connections. Figure \ref{fig:mpn_cc} shows BaseCNN-2-CC architecture which has two parallel paths. We insert a one-to-many connector (cross-connecting layer connecting one tensor to a given number of tensors) to expand the input image to parallel paths and add cross-connections after the $2^{nd}$, $4^{th}$ and $6^{th}$ convolutions and after the $1^{st}$ dense layer. Since a cross-connection-based routing layer contains only cross-connections and weighing coefficients, adding such a layer does not increase the effective depth of the network. Therefore we insert these layers into the BaseCNN multi-path network without replacing any forward layers.  
  
We double the filter size in each convolution to widen the BaseCNN, resulting in WideCNN. We also add more convolutional layers to the BaseCNN, which results in the DeepCNN architecture. To compare with an equivalent multi-path network which does not have intermediate routing, we build BaseCNN-X. Here, X stands for the  number  of  parallel  BaseCNNs  sharing  the  same  input and output (averaging). To compare with model ensembles, we use an ensemble of 3 BaseCNNs trained individually (Base Ensemble). The output of the Base Ensemble is computed by averaging the individual BaseCNN responses at inference. We also build an ensemble of BaseCNN, WideCNN and DeepCNN, referred to as All Ensemble. 

To compare our multi-path networks with equivalent SENets \citep{hu2017squeeze}, we add SE operations in convolutional layers of BaseCNN and DeepCNN, which results in SEBaseCNN and SEDeepCNN respectively. We replace the adaptive cross-connections in BaseCNN-2-CC with cross-stitching operations to build an equivalent two-path Cross-Stitch Network \cite{cross_stich}, Cr-Stitch2. We replace the cross-prediction operations in BaseCNN-2-CP with NDDR operations to build the equivalent two-path NDDR-CNN \cite{nddr-cnn} (NDDR-CNN2). In addition, we also build NDDR-CNN2-shortcut \cite{nddr-cnn}, which has shortcut connections in the convolutional part. NDDR-CNN2-shortcut network generalizes both cross-stitching operations and weighted skip connections in Sluice Networks.

First, we train these models in the CIFAR10 dataset for 200 epochs with a batch size of 128. We use Stochastic Gradient Descent (SGD) with a momentum of 0.9 and an initial learning rate of 0.1, which is decayed by a factor of 10 after 80 and 150 epochs. We augment the input images by random pixel shift in both directions with a maximum shift of 4 pixels and random horizontal flipping. Table \ref{tab:cnn} shows the results of this study. For each model, we report the best performance out of 3 trials.

\begin{table}[t]
	\caption{Ablation study of CNNs with CIFAR10 - Classification errors (\%). BaseCNNs with parallel paths and routing, at similar or less complexity, show superior performance to conventional widening, model ensembles, SENets, Cross-stitch networks and even conventional deepening. Considering the number of parameters utilized, adaptive cross-connections show the best performance. All networks are trained for 200 epochs. We further report our multi-path network performance after training for 350 epochs to set the benchmark (Column Error$\%^\dagger$). Among the compared networks, * denotes the performance stated in the respective paper.}
	\label{tab:cnn} 
	\begin{center}
	\renewcommand{\tabcolsep}{1.2mm}
	\small
	\begin{tabular}[width=\columnwidth]{@{}lccr@{}}
		\toprule
 	    Network             & Params (M) & Error\% & Error$\%^\dagger$ \\
		\midrule 
		 BaseCNN             &0.55 	&9.26  &\\
		 WideCNN            &1.67   &8.96 &\\
		 DeepCNN            &2.0    &8.49 &\\
         BaseCNN-3 			&1.5	&9.41 &\\
	     BaseCNN Ensemble    &1.66 	&7.87  &\\
		 All Ensemble        &4.27  	&6.9    &\\
		 SEBaseCNN		&0.58 	&8.99   &\\
         SEDeepCNN 		& 2.06 	&8.15	&\\
		 Cr-Stitch2          &1.11 	&7.89    &\\
          NDDR-CNN2          &0.96 	&7.81    &\\
          NDDR-CNN2-shortcut          &0.99 	&8.33    &\\
		 VGG16 \cite{vgg}    &14.9 	&6.98  &\\
		 Capsule Nets* \cite{sabour2017dynamic}       &8.2    &10.6  &\\
		 Highway Nets* \cite{srivastava2015highway, highway2} &2.3  &7.54 &\\ 
		 \midrule 
         \textbf{BaseCNN-2-CP}         &1.3 	    &  7.24     &\textbf{6.48}\\
		 \textbf{BaseCNN-3-CP}          &2.23 	&\textbf{6.63}   &\textbf{6.04} \\
		 \textbf{BaseCNN-4-CP}          &3.34 	&\textbf{6.45}   &\textbf{5.91}\\
		 \textbf{BaseCNN-2-CC}          &1.11 	&7.03    &\textbf{6.53}\\
		 \textbf{BaseCNN-3-CC}          &1.67 	&\textbf{6.51}    &\textbf{6.09}\\
		 \textbf{BaseCNN-4-CC}          &2.22 	&\textbf{6.55}   &\textbf{6.26}\\
		\bottomrule
	\end{tabular}
	\end{center}
\end{table}

Adding parallel paths to BaseCNN with our routing algorithms improves the performance of BaseCNN and also surpasses conventional widening. In this particular setting, BaseCNN with two paths, and our routing (BaseCNN-2-CP/CC) is sufficient to surpass the WideCNN, which has two times filters in each layer. Due to the quadratic increment of parameters with conventional widening, WideCNN carries nearly four times the parameters of BaseCNN, whereas having two parallel paths only doubles the number of parameters. Even with the routing overhead added, the total number of parameters of BaseCNN-2-CP is still significantly less than WideCNN, where BaseCNN-2-CC carries almost the same amount of parameters as two BaseCNNs due to cross-connection-based routing, adding a minimal amount of routing overhead. 

BaseCNN-3-CP and -CC, with a clear margin, show superior performance to BaseCNN-3, which does not have intermediate routing. Also, BaseCNN-3-CP and -CC outperform the ensemble of 3 BaseCNNs, and even the ensemble of BaseCNN, WideCNN and DeepCNN. This indicates that the improvement of our multi-path networks is not merely due to the widened nature, but also due to the adaptive routing mechanisms. BaseCNN-2-CP/CC even surpasses the DeepCNN, whose total number of parameters is more than three times the parameters in the BaseCNN. Finally, our multi-path networks surpass the VGG16 \citep{vgg}, which consists of many parameters along the depth and the width.

BaseCNN-2-CP/CC surpasses the cross-stitch network (Cr-Stitch2) and NDDR-CNNs with two paths (NDDR-CNN2 \& NDDR-CNN2-shortcut), proving that adaptive cross-routing is more suitable for learning a task while handling the diversity in input rather than independently learned cross-connecting coefficients. BaseCNN-2-CP/CC further surpasses the SE Nets built based on the WideCNN and DeepCNN, showing the effectiveness of utilizing parallel paths over the re-calibration of a single path. Among the other methods for rich layer-wise feature extraction or adaptive feature extraction, ours surpass Highway networks \citep{srivastava2015highway} and Capsule Networks \citep{sabour2017dynamic} at similar or less complexity. 

Adding a parallel path to BaseCNN (BaseCNN-2-CP/CC) significantly improves the BaseCNN performance with CIFAR10. However, the performance gain is not that significant with the addition of the third parallel path (BaseCNN-3-CP/CC). Adding the fourth path (BaseCNN-4-CP/CC) gives little or no improvement. Therefore, it is essential to carefully design the number of parallel paths according to the dataset to get the best performance for the number of parameters utilized. However, this phenomenon is common to all deepening \citep{resnet, preact-resnet} and widening \citep{wideresnet, resnetxt} techniques. 

The multi-path networks with cross-connections (BaseCNN-X-CC) use significantly less number of parameters compared to the networks with cross-prediction-based routing (BaseCNN-X-CP), which is more prominent with the increased number of parallel paths. This is because adaptive cross-connections drastically reduce the routing overhead by eliminating the cross-convolutions or cross-dense operations in cross-prediction-based routing. Cross-connection-based routing also performs similarly to cross-prediction-based routing, yielding better performance with respect to the model complexity. We further set the benchmark for CNN-based multi-path networks: We re-train our multi-path nets in the previous setting but for 350 epochs, where the learning rate decayed after 150 and 250 epochs. The benchmark values are shown in the final column of Table \ref{tab:cnn}.

\subsection{Residual Networks with Parallel Paths}
\label{ss:multi_resnet}

\begin{table}[t]
	\caption{Comparison of ResNets. ResNet20-3 outperforms ResNet110. ResNet20-3/4 and ResNet32-3/4 show on-par or superior performance to existing adaptive architectures which are mostly based on ResNet110.}
	\label{tab:resnet}
	\begin{center}
	\renewcommand{\tabcolsep}{0.7 mm}
	\small
	\begin{tabular}[width=\columnwidth]{@{}lccr@{}}
		\toprule
 	    Network             & Params (M) & CIFAR10 &CIFAR100   \\
		\midrule 
		 ResNet20 \cite{resnet}   &0.27 	&8.75    &-\\
		 ResNet110                &1.7    &6.61   &26.88\\
		 ResNet164                &2.5   &5.93   &25.16\\
		 WRN-40-2 \cite{wideresnet} &2.2   & 5.33    &26.04 \\
		 HyperWRN40-2 \cite{ha2016hypernetworks_arxiv}  & 0.15   &7.23 &-\\
		 SEResNet110 \cite{hu2017squeeze}             &1.7   &5.21    &\textbf{23.85}\\
		 BlockDrop \cite{blockdrop}                   &1.7   &6.4  &26.3\\
	     ConvNet-AIG \cite{convnet-aig}               &1.78   &5.76   &-\\
	     ConvNet-AIG all \cite{convnet-aig}            &1.78   &5.14   &-\\
        SkipNet  \cite{wang2018skipnet}           &1.7   &6.4   &28.79\\
         \midrule
	     \textbf{ResNet20-2-CP}       &0.59 	   &5.86    &27.7\\
		 \textbf{ResNet20-3-CP}       &0.92     &\textbf{4.99}    &\textbf{25.13}\\
		 \textbf{ResNet20-4-CP}       &1.29      &\textbf{4.81}   &\textbf{23.82}\\
         \midrule
	     \textbf{ResNet20-2-CC}      &0.55 	  &5.5    &27.36\\
		 \textbf{ResNet20-3-CC}       &0.82     &5.18   &25.76\\
		 \textbf{ResNet20-4-CC}       &1.1      &\textbf{4.96}   &\textbf{24.81}\\
		 \textbf{ResNet32-2-CC}       &0.94 	 &5.14    &25.96\\
		 \textbf{ResNet32-3-CC}       &1.41     &\textbf{4.96}   &\textbf{24.51}\\
		 \textbf{ResNet32-4-CC}       &1.88     &\textbf{4.59}   &\textbf{23.52}\\
	     \bottomrule
	\end{tabular}
	\end{center}
\end{table}

\begin{figure*}[t]
\begin{center}
    \begin{subfigure}{0.49\linewidth}
		\includegraphics[width=\linewidth]{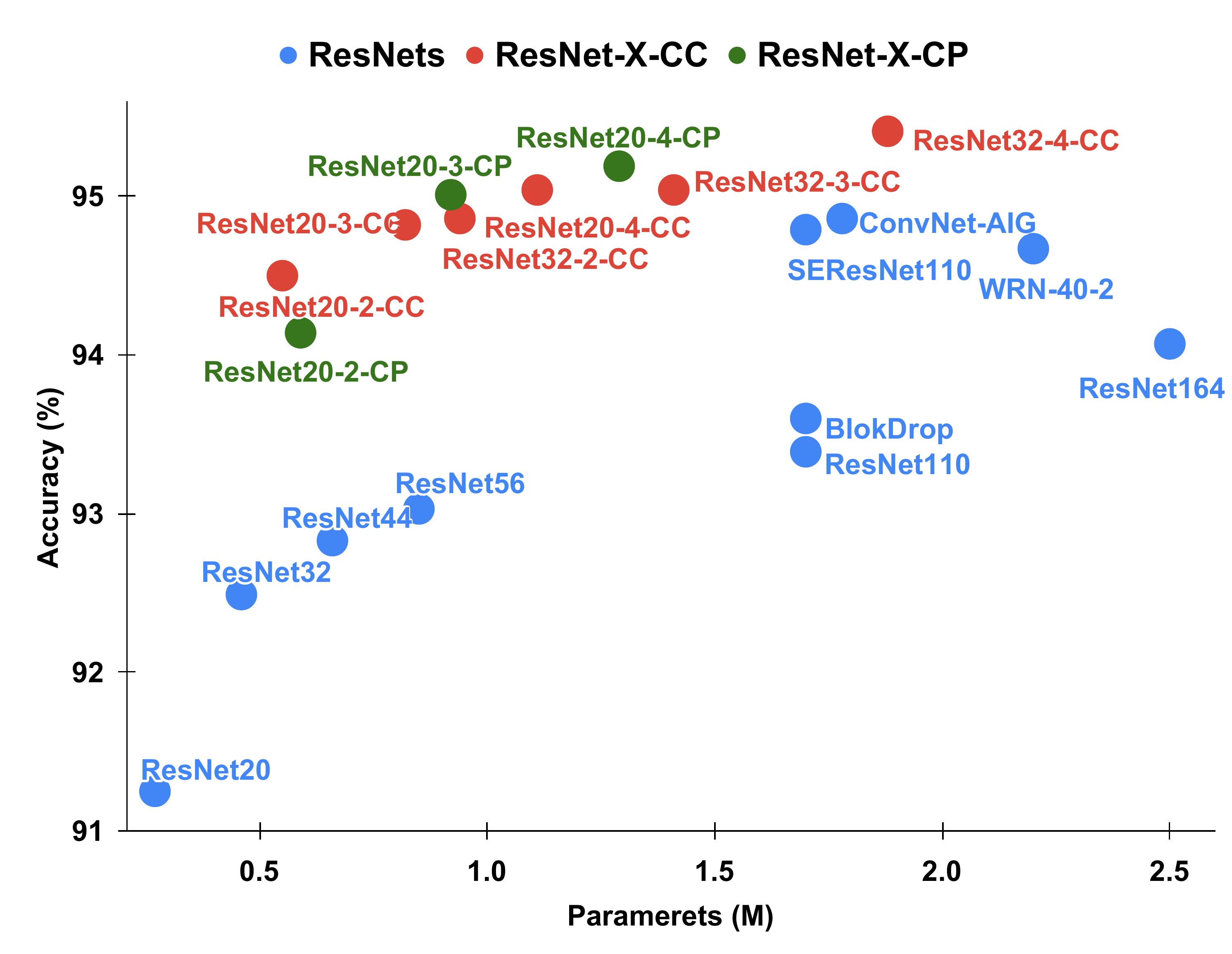} 
		\caption{CIFAR10}
		\label{fig:cifar10}
	\end{subfigure}
    \begin{subfigure}{0.49\linewidth}
		\includegraphics[width=\linewidth]{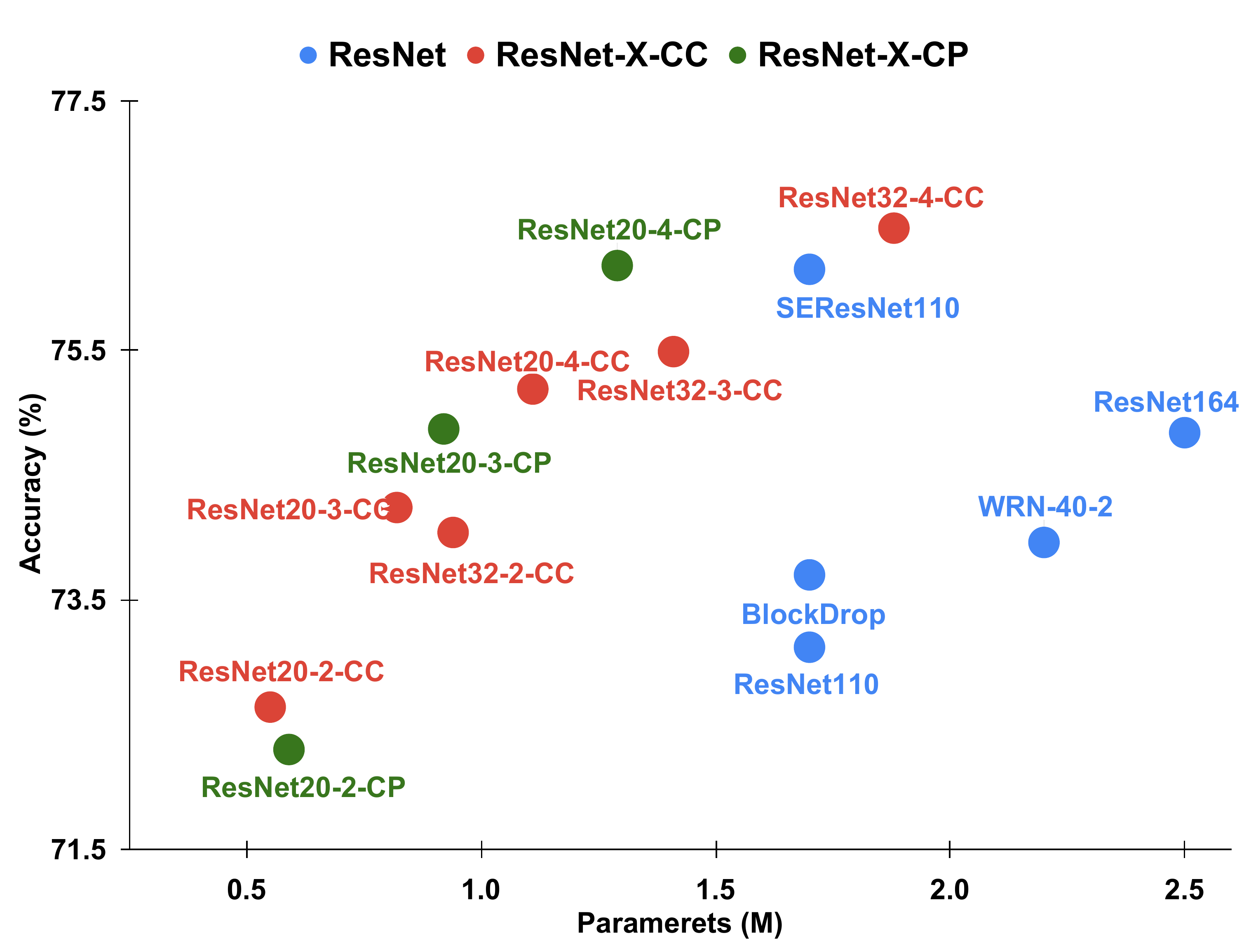} 
		\caption{CIFAR100}
		\label{fig:cifar100}
	\end{subfigure}
\end{center}
\caption{ResNet performance (accuracy) with CIFAR10 and CIFAR100, along with the number of parameters in millions. Blue circles correspond to conventional ResNets and ResNet-based adaptive networks. Green circles show multi-path ResNets with cross-prediction-based routing. Red circles show multi-path ResNets with cross-connection-based routing. Our multi-path networks yield the best performance w.r.t the network complexity. Our networks (red and green) flocking to the top-left show their superior performance with fewer parameters, in general.}
\label{fig:cifar}
\end{figure*}

Next, we extend the residual networks (ResNets) \citep{resnet} with parallel paths and our routing schemes. First, we add parallel paths to the ResNet variants (ResNet20, ResNet32, etc.) designed to learn from small-scale datasets. In these models, an initial convolution is followed by three sequential stacks, where, in each stack, several residual blocks (In ResNet20, three residual blocks in each stack) are employed. Each stack starts with a strided residual block, resulting in down-sampled feature maps. The network terminates with a global average pooling layer, followed by the final dense layer, which outputs the class probabilities. 

We build parallel-path ResNets with cross-prediction-based routing (ResNet-X-CP) as following. First, we replace the initial convolutional layer with a convolutional one-to-many routing layer. Then we add two more routing layers before the $2^{nd}$ and $3^{rd}$ stacks. Finally, the parallel dense layer outputs are averaged to produce the output. This design adds two more layers to the effective depth. To build parallel-path ResNets with cross-connection-based routing (ResNet-X-CC), we add one-to-many connector after the initial convolution and three cross-connection-based routers after the $1^{st}$, $2^{nd}$ and $3^{rd}$ stacks. Since these cross-connections  do not contain convolutions, this design preserves the original depth of the network.

To train ResNet-based variants with CIFAR10 and CIFAR100 \citep{cifar100} datasets, we use a similar setting to the previous study. We use a batch size of 64 and train our models for 350 epochs, where the learning rate decays after 150 and 250 epochs. For each model, we conduct three trials and report the best performance. Table \ref{tab:resnet} shows the recorded classification errors of our models and the reported errors of conventional ResNets and ResNet-based adaptive feature extractors. 

ResNet20, with three paths, and our routing algorithms (ResNet20-3-CP/CC), surpasses the WideResNet40-2 (WRN-40-2), which has a depth of 40 layers and two times filters in each convolutional layer. The Hyper Network \citep{ha2016hypernetworks} built on top of WideResNet-40-2 (HyperWRN40-2) shows an inferior performance to the original WRN-40-2,  although it uses a few numbers of parameters. With CIFAR10, ResNet20 with two paths surpasses ResNet110, and with CIFAR100, ResNet20 with three parallel paths surpasses ResNet110. This is impressive, as compared to ResNet110, ResNet20 is very shallow, and even with parallel paths added (2/3/4), the total number of parameters is still less than ResNet110. 

Furthermore, ResNet-based multi-path networks surpass existing adaptive feature extraction methods built on ResNet110. BlockDrop \cite{blockdrop} and SkipNet \cite{wang2018skipnet} architectures, built on ResNet110, show inferior performance to all our multi-path networks with CIFAR10. With CIFAR100, BlockDrop only shows better performance to ResNet20-2-CP/CC where SkipNet shows inferior performance to all our mult-path networks. ResNet20-3/4-CP, ResNet20-4-CC, and ResNet32-3/4-CC show superior performance to the ConvNet-AIG \cite{convnet-aig}, based on ResNet110. All our multi-path networks except ResNet20-2-CC/CP surpass the SENet \cite{hu2017squeeze}, built using ResNet110 with identity mappings \citep{preact-resnet} with CIFAR10. With CIFAR100, ResNet20-4-CP shows on-par performance with SEResNet110, and ResNet32-4-CC surpasses its performance. Among our multi-path nets, all the networks other than ResNet32-4-CC have less number of parameters than ResNet110-based networks. 

Figure \ref{fig:cifar} plots the accuracies of the compared networks in CIFAR along with the number of parameters utilized. These plots clearly illustrate that our multi-path networks show the best utility of the network for the used number of parameters. Multi-path ResNets with cross-prediction-based routing give the best performance for a given depth. However, we prefer cross-connection-based multi-path ResNets due to the less complex routing algorithm, which adds significantly less routing overhead to the widening.

\subsection{Multi-path ResNets on ILSVRC2012}
\label{ss:imagenet}

\begin{table}[t]
	\caption{Single-crop and 10-crop validation error (\%) in ILSVRC2012 dataset. ResNet18-2, with two paths, comfortably outperforms ResNet18 and shows on-par performance with ResNet34. It also surpasses the WideResNet18, which has 1.5 times as filters in each layer. In the subset of ILSVRC2012, which contains the first 100 classes, ResNet50-2-CC, with similar or fewer model parameters, outperforms WideResNet and ResNeXt counterparts and even the twice deep ResNet101. * denotes reproduced results}
	\label{tab:classi_img}
	\begin{center}
	\renewcommand{\tabcolsep}{0.8mm}
	\small
	\begin{tabular}[width=\columnwidth]{@{}lccccr@{}}
		\toprule
        Network  & Params  &\multicolumn{2}{c}{Single-Crop} & \multicolumn{2}{c}{10-Crop} \\
 	               & & Top-1 & Top-5 & Top-1 & Top-5  \\
		 \midrule 
		 \multicolumn{6}{c}{Full Dataset} \\
		 \midrule
		 ResNet18 \cite{wideresnet, fbresnet}       & 11.7M   &30.4 	&10.93    &28.22    &9.42 \\
		 ResNet34 \cite{wideresnet, resnet}   & 21.8M  &26.77 	&8.77    &24.52    &7.46 \\
		 WRN-18-1.5 \cite{wideresnet}     & 25.9M &27.06 & 9.0 & &   \\
		 \textbf{ResNet18-2-CC}                & 23.4M &\textbf{26.48} 	&\textbf{8.6}     &\textbf{24.5}    &\textbf{7.34}\\
		 \midrule 
		 \multicolumn{6}{c}{Subset of first 100 classes} \\
		 \midrule
		 ResNet50*           &23.71M &20.46 &4.96 &19.26 &4.72 \\
		 ResNet101*          &42.7M &19.16 &4.58 &17.78 &4.44\\
		 WideResNet50-2* \cite{wideresnet}     &62.0M &19.82 &5.02 &18.62 &4.76 \\
		 ResNeXt50-2-64* \cite{resnetxt}     &47.5M &20.26 &5.06 &19.0 &4.84   \\
		 \textbf{ResNet50-2-CC}  &47.5M &\textbf{18.64} &\textbf{4.34} &\textbf{17.62} &\textbf{4.0} \\
		 \bottomrule
	\end{tabular}
	\end{center}
\end{table}

Here, we further evaluate our multi-path ResNets in the ILSVRC 2012 Dataset \cite{deng2009imagenet, ILSVRC15}. To train with this dataset, we expand the residual networks originally designed to learn in the ImageNet dataset \citep{resnet} with parallel paths. These residual networks share a similar setting to the thin residual networks designed to learn from CIFAR. These have an initial 7$\times$7 convolution with a stride of 2 followed by a max-pooling operation. After that, four sequential stacks of residual blocks are employed, where each stack contains a pre-defined number of residual blocks sharing the same feature map size. Each stack's first residual operation starts with a strided convolution which downsamples the feature maps by a factor of 2. The final residual block's response is fed to a global average pooling operation and the final fully connected layer, which outputs the class response.

The cross-connection-based routing is less complex, uses very little overhead, and still gives reasonably similar results to cross-prediction-based routing. Thus, we only use cross-connection-based routing in expanding these models to parallel paths. In particular, after the initial convolution and max-pooling, we insert a one-to-many connector, which expands the network to parallel paths and insert cross-connection-based routing layers after each stack containing residual blocks of certain feature map size. Finally, we average the final layer parallel dense predictions. 

We expand ResNet18 with two parallel paths and cross-connection-based routing (ResNet18-2-CC) and train in the dataset for 120 epochs with a batch size of 256. We use SGD optimizer with a momentum of 0.9 and an initial learning rate of 0.1, which is decayed by a factor of 10 after every 30 epochs. We use standard data augmentation of re-scaling to 256$\times$256, taking random crops of 224$\times$224, and randomly flipping in the horizontal axis. To further evaluate deeper models with parallel paths, we use a subset of the ILSVRC dataset, which only contains the first 100 classes. This subset contains 130k training images and 5k validation images. To learn in this subset, we expand ResNet50 with two paths and cross-connection-based routing (ResNet50-2-CC). We use a similar training setup as in the full dataset, except that the models are trained for 90 epochs. To compare with ResNet50-2-CC in this subset, we train ResNet50 and WideResNet50-2, which has two times filters in each layer, and ResNeXt50-2-64, which has two parallel operations in each layer, and ResNet101. 

Table \ref{tab:classi_img} shows the results of this study. ResNet18, with two parallel paths and cross-connections, in the ILSVRC 2012, comfortably surpasses the performance of the single path ResNet18 and shows on-par performance to ResNet34. It also surpasses the performance of WideResNet18 with 1.5 times convolutional filters in each layer which still has more parameters than ResNet18-2-CC. In the subset, ResNet50-2-CC surpasses its single path baseline (ResNet50) and both WideResNet50-2 and ResNeXt50-2-64, confirming the superiority of our approach to existing widening at similar complexity. ResNet50-2-CC even shows slightly better results than ResNet101, which is twice deep.

Overall, these experiments validate that our multi-path networks, along with the adaptive routing algorithms, show efficient usage of the resources in each layer. Due to this efficient use of layer resources, our multi-path networks, at similar or less complexity, show superior performance to conventional widening and other methods for rich layer-wise feature extraction and even conventional deepening.

\section{Visualization of Multi-path Learning}
\label{se:visualization}

\begin{figure*}[t]
	\begin{center}
	\includegraphics[width=0.9\linewidth]{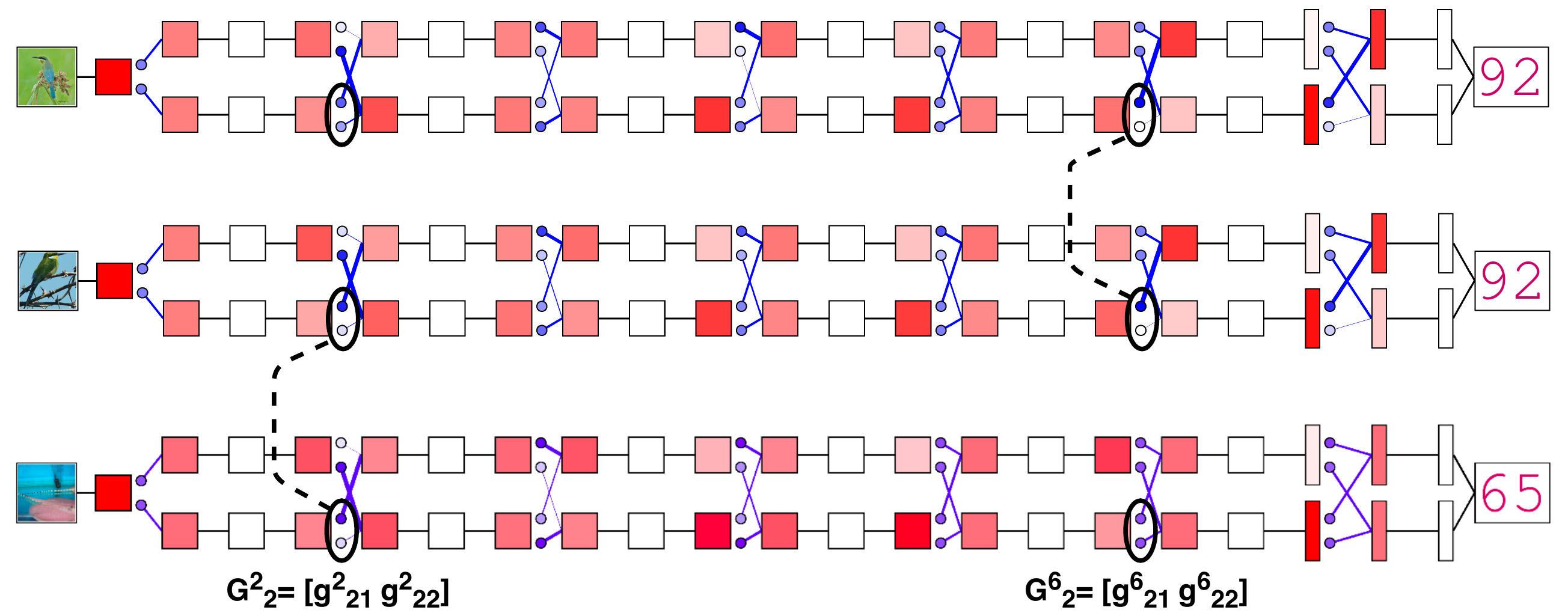}
	\end{center}
	\caption{Route visualizations through cross-connections of VGG13-2-CC for the three images in Fig. \ref{se:intro}. The top routing diagram relates to Image \ref{fig:humming1} (Hummingbird in green background), the middle diagram to Image \ref{fig:humming2} (Hummingbird in blue background), and the bottom diagram to Image \ref{fig:madu} (Electric ray in water). In each cross-connecting layer, the relative strengths of input and output tensors are shown in red intensities, and the gate strengths are shown in blue intensities and connection thicknesses. The gating vector $G^2_2$ that lies in shallow layers, shows similar gating patterns for Image \ref{fig:humming2} and Image \ref{fig:madu}, which belong to two different classes but share similar background colors. However, the gating vector $G^6_2$, withing deeper layers, shows similar gating patterns for the two hummingbird images, \ref{fig:humming1} and \ref{fig:humming2}. The resource allocation in each routing layer is sensitive to the features represented by that depth.}
	\label{fig:routes}
\end{figure*}

In this section, we use several visualization techniques to study the gating patterns of the cross-connection-based routing scheme. For this purpose, we use a VGG13 \citep{vgg} network with half the filters (32, 64, 128, 256) in each convolutional layer and 256 nodes in each dense layer. We join two such networks through cross-connections to build VGG13-2-CC, where the routing layers are added after each pooling operation and after the first dense layer, following a similar pattern to the multi-path networks in Section~\ref{se:cc}. We train this network with a subset of the ILSVRC2012, which contains the first 100 classes.

First, we visualize the routing patterns of this trained network and show the differences in gating patterns observed in layers at varying depths of the network. We maximize a set of selected gating neurons to understand these gating patterns further. We show images from the validation datasets that mostly activate those neurons and further synthesize randomly initialized images that maximize those neurons. Also, we plot the gate activations of selected classes to understand the class-wise gate activation. Finally, we plot weight histograms of the two parallel paths at selected layers to demonstrate that each path can learn distinct information. 

\subsection{Visualization of Routing}
\label{ss:vis_routes}

We visualize the routing flow through cross-connections of the trained 2-path network to understand the gating patterns. Figure \ref{fig:routes} shows such visualizations for the three images depicted in Figure \ref{fig:intro}. For each cross-connection-based routing layer with two parallel inputs, two parallel outputs, and gates that weigh the connections, we plot the relative activation strengths of input and output tensors and the gate strengths. We calculate the relative activation strength of a tensor by taking the average activation value of that tensor and normalizing it by all such values of the parallel tensors of that layer. We map these relative activation strengths to red intensities and use these colors to color each box representing the particular tensor. The softmax gate values computed by each input are directly mapped to blue intensities and thickness values which are then used to color the circles denoting each gate and edges denoting each weighted connection, respectively. We denote the stacks of conventional forward layers by uncolored boxes. They contain sequential convolutions or dense operations which run in parallel, but no cross-operations are performed.

Let $G^l_i$ ([$g^l_{i1}$, $g^l_{i2}$]) be the gating vector computed by the $i^{th}$ input tensor of the $l^{th}$ cross-connecting layer. In these routing plots, we pay attention to the gating vectors $G^2_2$ ([$g^2_{21}$, $g^2_{22}$]) and $G^6_2$ ([$g^6_{21}$, $g^6_{22}$]). $G^2_2$, lying within the network's initial layers, shows similar gating patterns to image \ref{fig:humming2} and image \ref{fig:madu} (maximized $g^2_{21}$), although they belong to entirely different classes. At the same time, $G^2_2$ shows different gating patterns to image \ref{fig:humming1} and image \ref{fig:humming2}, although they are both hummingbirds. However, $G^6_2$, lying within a deeper layer of the network, shows similar gating patterns to the two hummingbird images (maximized $g^6_{21}$), while the gating pattern for the electric eel is significantly different. These visualizations show that a gating layer's behavior depends on the features captured in the corresponding network depth, and, based on the features at different depths of the network, the gating behavior changes. Thus, it is vital to have routing layers throughout the depth of the network. To further understand the basis of this behaviour, we next explore which features maximize each gate.  

\subsection{What Maximizes Gates?}
\label{ss:act_max}

\begin{figure*}[t]
	\begin{center}
	\begin{subfigure}{0.7\linewidth}
		\includegraphics[width=\linewidth]{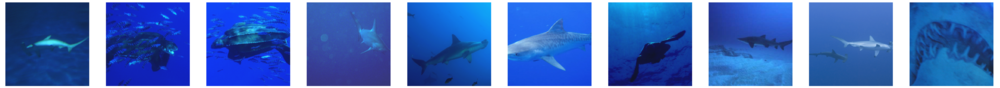}
		\includegraphics[width=\linewidth]{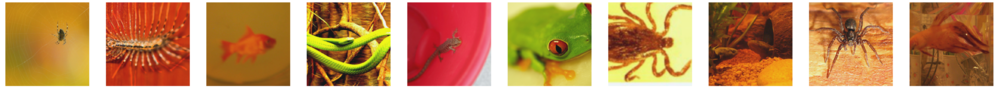}
		\caption{$g^2_{21}$}
		\label{fig:g^2_21}
	\end{subfigure}
	\begin{subfigure}{0.22\linewidth}
			\includegraphics[width=\linewidth]{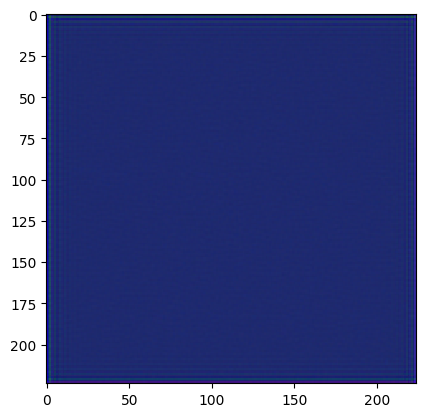}
	\end{subfigure}
	\begin{subfigure}{0.7\linewidth}
		\includegraphics[width=\linewidth]{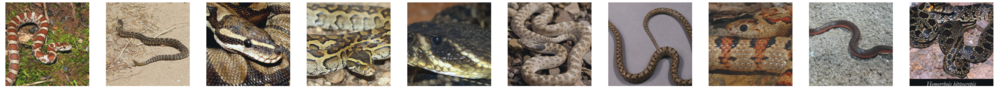}
		\includegraphics[width=\linewidth]{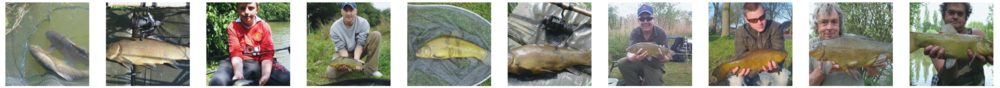}
		\caption{$g^6_{11}$}
	\end{subfigure}
	\begin{subfigure}{0.22\linewidth}
			\includegraphics[width=\linewidth]{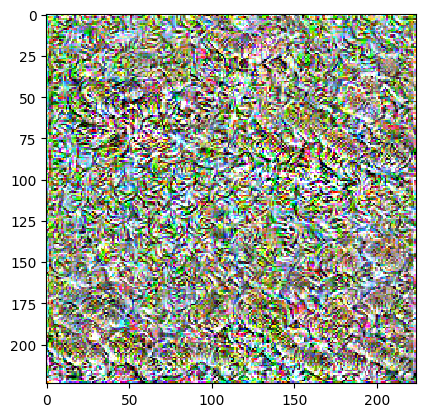}
	\end{subfigure}
	\begin{subfigure}{0.7\linewidth}
		\includegraphics[width=\linewidth]{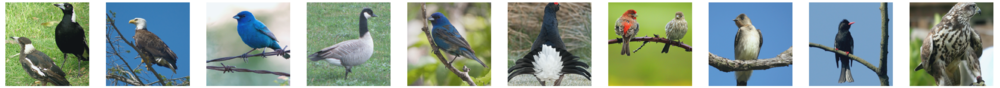}
		\includegraphics[width=\linewidth]{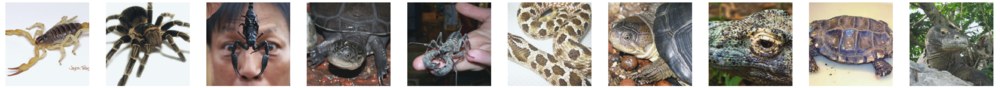}
		\caption{$g^6_{21}$}
		\label{fig:g^6_21}
	\end{subfigure}
	\begin{subfigure}{0.22\linewidth}
			\includegraphics[width=\linewidth]{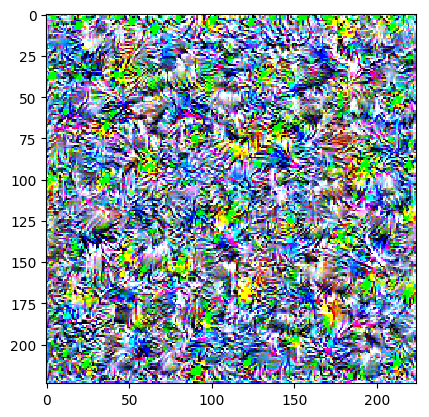}
	\end{subfigure}
	\begin{subfigure}{0.7\linewidth}
		\includegraphics[width=\linewidth]{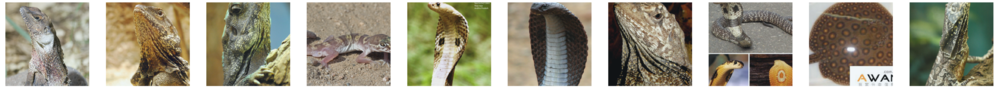}
		\includegraphics[width=\linewidth]{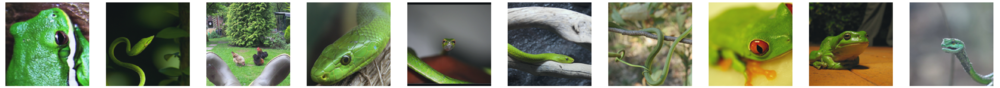}
		\caption{$g^7_{11}$}
	\end{subfigure}
	\begin{subfigure}{0.22\linewidth}
			\includegraphics[width=\linewidth]{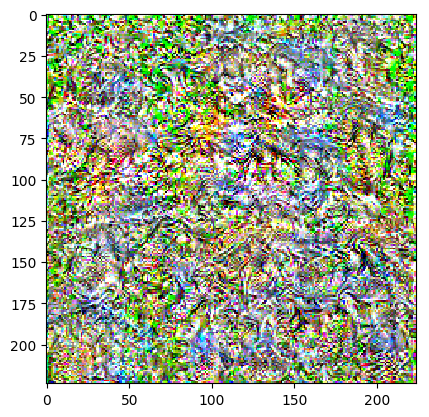}
	\end{subfigure}
	\end{center}
	\caption{Maximization of selected gates: Each subfigure, corresponding to a particular gate, shows the ten images with the highest gate activation (top left), the ten images with the lowest gate activation (bottom left), and the synthesized image such that the gate neuron is maximized. $g^2_{21}$, which is within initial layers, is maximized for blue while the other gates which lie within deeper layers get triggered for more abstract features such as snake body patterns ($g^6_{11}$), bird patterns ($g^6_{21}$) and raised upper body patterns ($g^7_{11}$).}
	\label{fig:act_max}
\end{figure*}

\begin{figure*}[t]
	\begin{center}
	\begin{subfigure}[b]{0.45\linewidth}
		\input{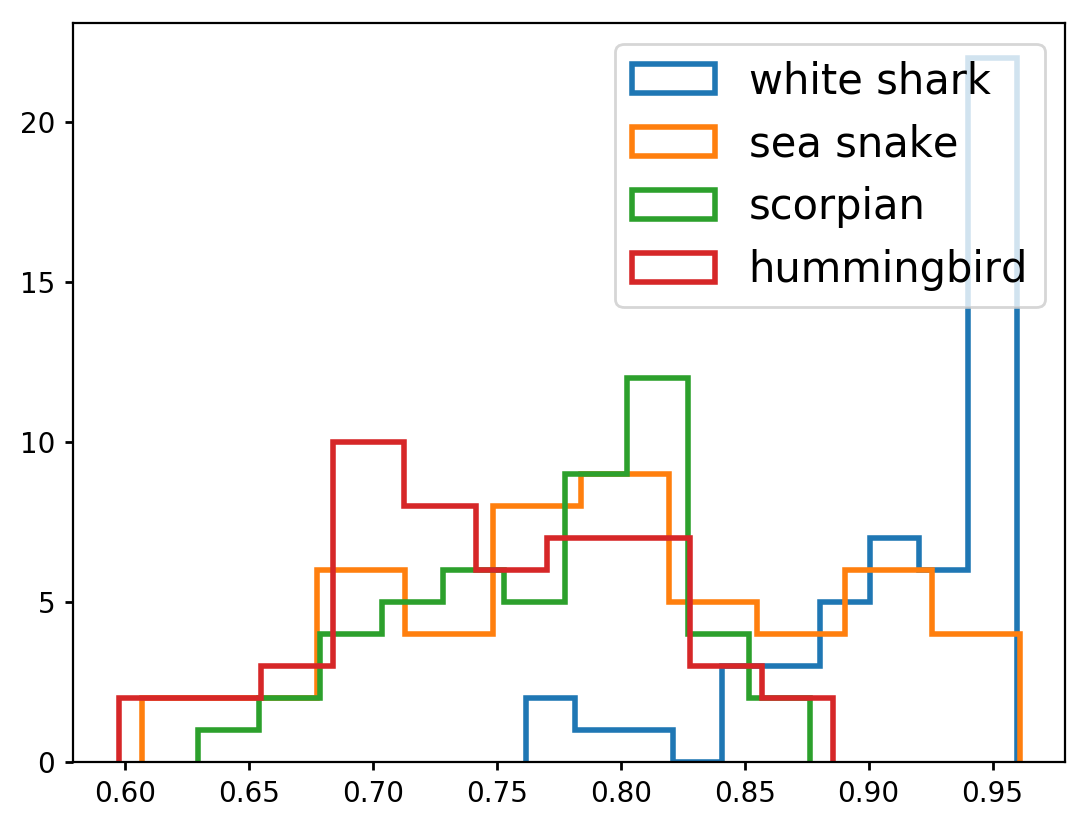}
		\caption{$g^2_{21}$}
	\end{subfigure}
	\begin{subfigure}[b]{0.45\linewidth}
		\input{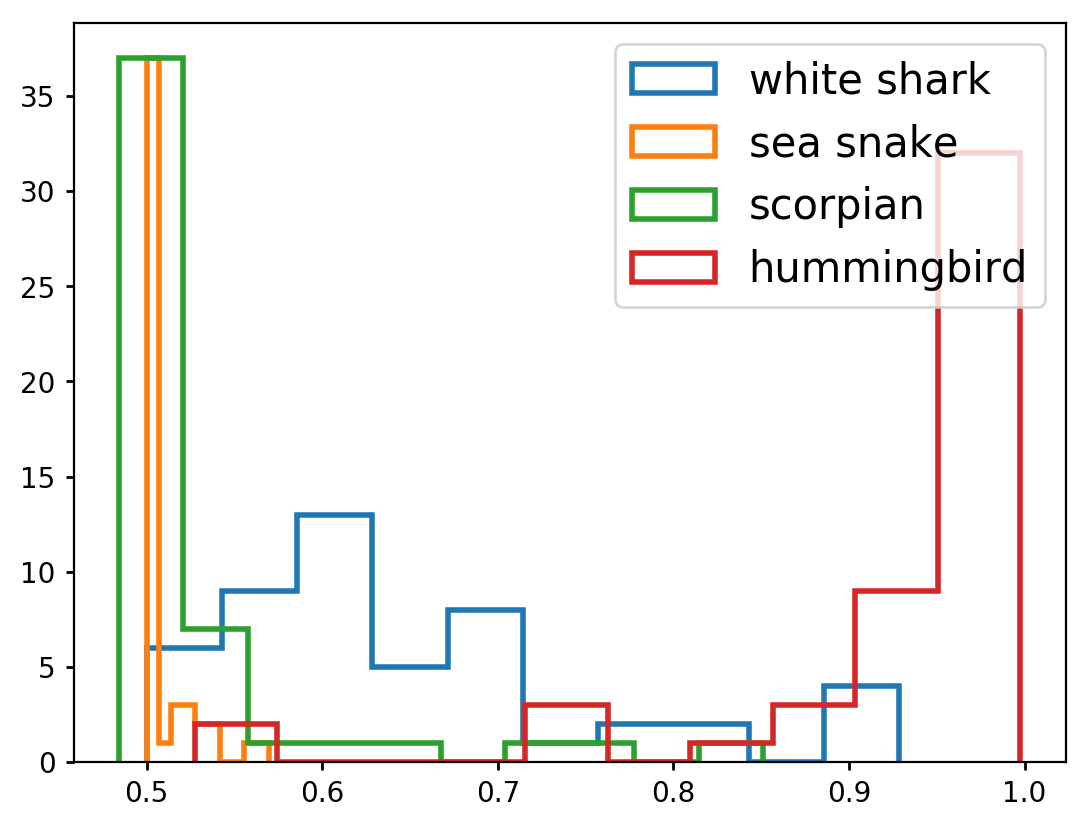}
		\caption{$g^6_{21}$}
	\end{subfigure}
	\end{center}
	\caption{Gate activation histograms for chosen four classes in ILSVRC 2012 validation set. The horizontal axis represents the gate activation value (0-1), and the vertical axis denotes the frequencies of bins. White shark, with mostly blue seawater as the background, has overall high activation for $g^2_{21}$ while other classes are evenly distributed. Hummingbird images mostly activate $g^6_{21}$ while other classes, agreeing less with bird patterns, show less $g^6_{21}$ activation. If the triggering pattern of any gate is commonly found in a certain class, those class members get similar gating in that layer.}
	\label{fig:gate_hist}
\end{figure*}

To understand the type of features (context) which maximizes a particular gate and, hence, to describe the gating patterns above, we plot the images in the validation set that maximally and minimally activate a particular gating neuron. Also, we freeze the trained network and synthesize the network input such that the particular gating neuron (before softmax activation) is maximized. This neuron maximization is similar to the gradient ascent process introduced in Simonyan \etal \cite{act_max}. We choose four gating vectors, $G^2_2$, $G^6_1$, $G^6_2$, and $G^7_1$, for this visualization.  $G^2_2$ and $G^6_2$ are the gate vectors we discussed in the previous visualization. Since one gating neuron is inversely related to the other gating neuron among a gating pair due to the softmax activation, maximizing one gate minimizes the other. Therefore we choose only one gating neuron from each pair for visualization--gating neurons $g^2_{21}$, $g^6_{11}$, $g^6_{21}$, and $g^7_{11}$ from each gating vector, respectively.

Figure \ref{fig:act_max} shows the output of this visualization. In each subfigure, the ten images which give the maximum gate activation are plotted in the top left, and the ten images which give the minimum activation are plotted in the bottom left. The synthesized image that maximizes the gate neuron is shown towards the right. Gate $g^2_{21}$ (Fig \ref{fig:g^2_21}), within initial layers, is maximized for the overall color of blue, which is a fairly low-level detail. However, all other gates within the deeper layers get maximized for rather complex patterns. $g^6_{11}$ gets maximized for body patterns of snakes, $g^6_{21}$ is maximized with bird poses and patterns, and $g^7_{11}$ is triggered best by animal poses with the raised thorax. The synthesized image's gate-maximizing patterns in each case agree with the top ten activated images.

Based on the maximization patterns of $g^2_{21}$ and $g^6_{21}$, we can interpret the gating behavior in the routing visualizations shown in Section \ref{ss:vis_routes}. Gate $g^2_{21}$ is maximally activated for blue; therefore, with the backgrounds highly composed of blue, Image \ref{fig:humming2} and Image \ref{fig:madu} maximize $g^2_{21}$ although they belong to different classes. Meanwhile, Image \ref{fig:humming1}, with green background, shows a lower $g^2_{21}$ activation although it belongs to the same class as Image \ref{fig:humming2}. Gate $g^6_{21}$, within deeper layers, gets maximized for bird posses and patterns. As a result, the two hummingbird images (Image \ref{fig:humming1} and Image \ref{fig:humming2}) maximize this gate while the electric eel (Image \ref{fig:madu}) shows a lower activation. This behavior highlights that the image context, which is related to the task, is distributed along with the depth of the trained network. Since resource allocation in different stages of depth varies depending on the level of context represented in that depth, it is vital to have routing layers within the network per segment of layers.

\subsection{Class-Wise Gating Patterns}
\label{ss:gate_hist}

The resource allocation in each layer of our multi-path networks depends on the nature of the feature maps in that particular depth. Therefore, to investigate any class influence on gating patterns, we plot the gate response of selected classes for gates $g^2_{21}$ and $g^6_{21}$. We choose four classes for this purpose, namely, white shark, sea snake, scorpion, and hummingbird, and record the gate response for all images belonging to each class in the ILSVRC 2012 validation set. Figure \ref{fig:gate_hist} summarizes the gate activation histograms for these four classes. 

The class white shark, having blue sea water as dominant detail in most cases, shows overall high activation for $g^2_{21}$. The other classes show an even distribution of $g^2_{21}$ since those classes contain instances that may or may not contain dominant blue. Also, the class hummingbird, with bird posses and patterns,  shows overall high activation for $g^6_{21}$, which triggers bird patterns. However, the other classes show overall less $g^6_{21}$ activations since they hardly agree on bird patterns. These observations reveal that the image context which matters to the gating---hence, resource allocation in each layer---is an intricate detail that expands beyond just the class. However, if the triggering pattern for a particular gate is mostly a part of a specific class, most class members shall show similar activations of that gate.

\subsection{Weights of Parallel Computations}
\label{ss:weight_hist}

\begin{figure*}[t]
	\begin{center}
	\begin{subfigure}[b]{0.4\linewidth}
		\input{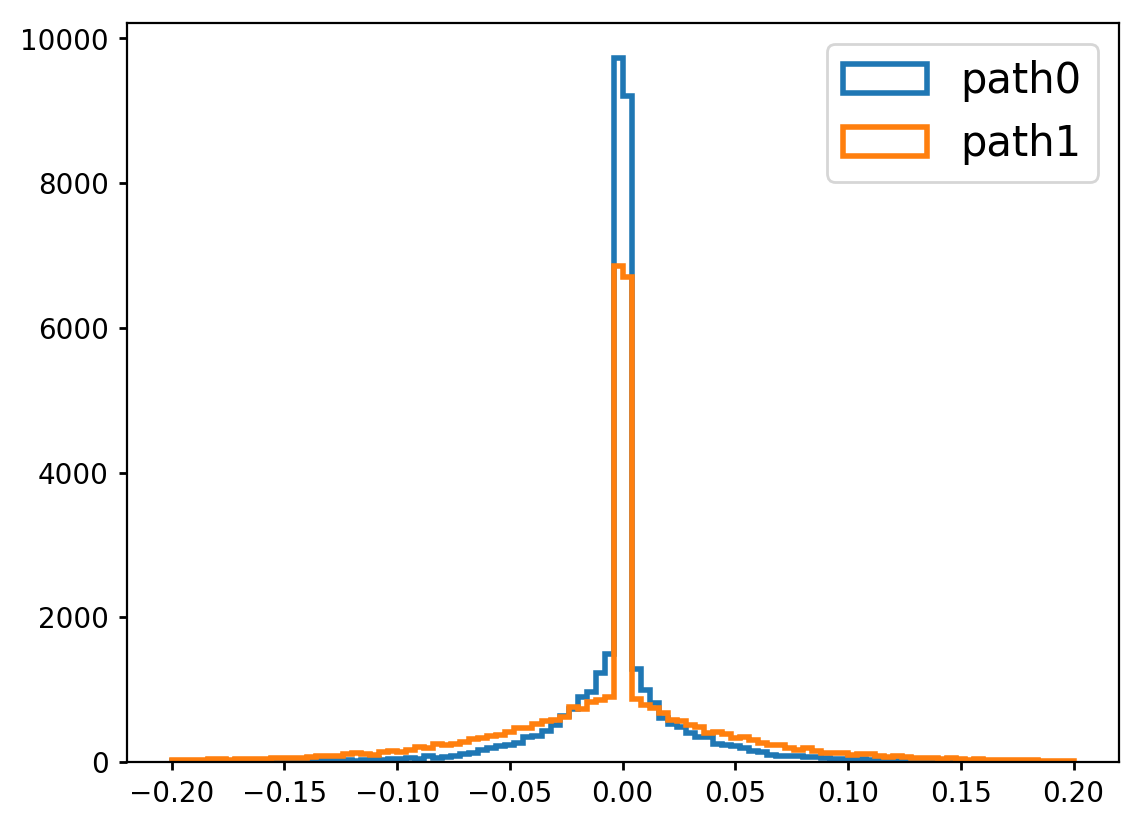}
		\caption{layer 4}
	\end{subfigure}
	\begin{subfigure}[b]{0.4\linewidth}
		\input{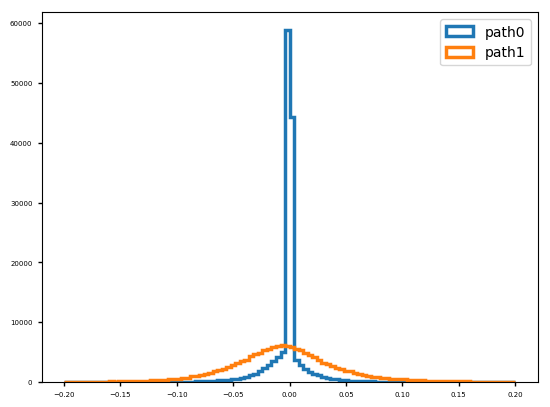}
		\caption{layer 6}
	\end{subfigure}
	\begin{subfigure}[b]{0.4\linewidth}
		\input{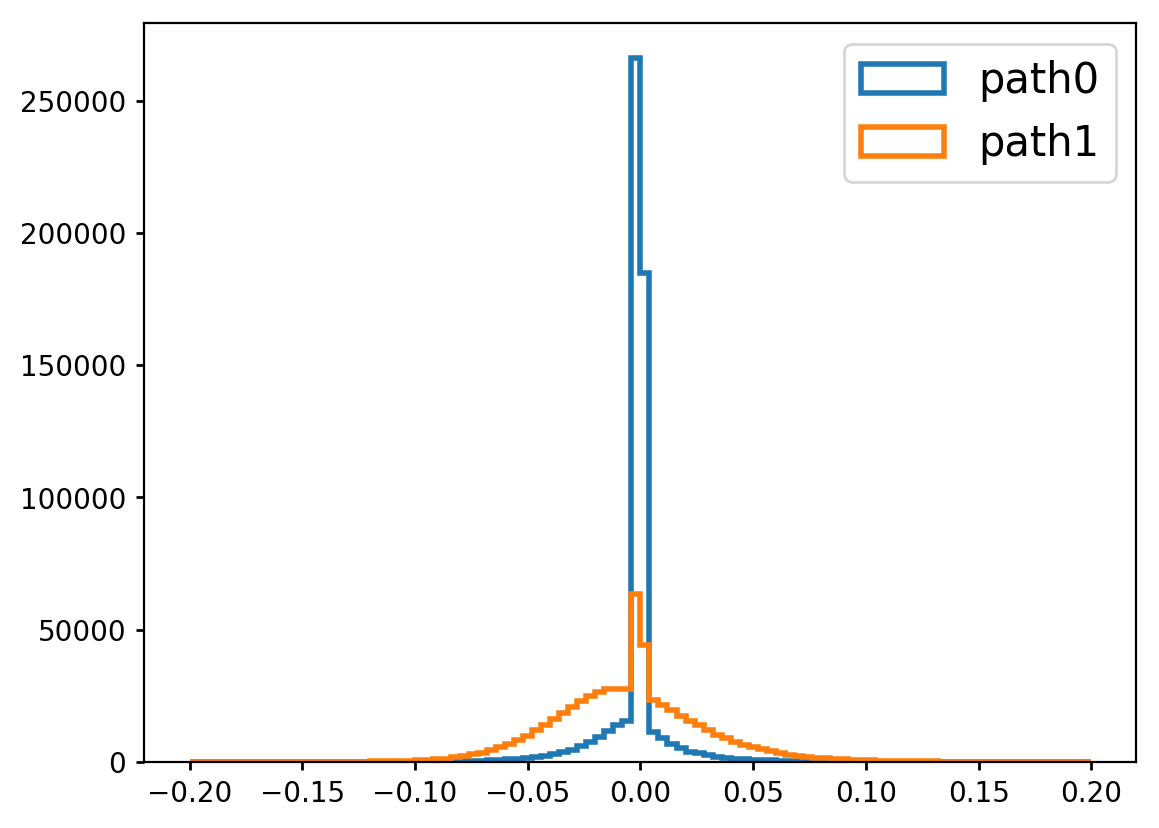}
		\caption{layer 8}
	\end{subfigure}
	\begin{subfigure}[b]{0.4\linewidth}
		\input{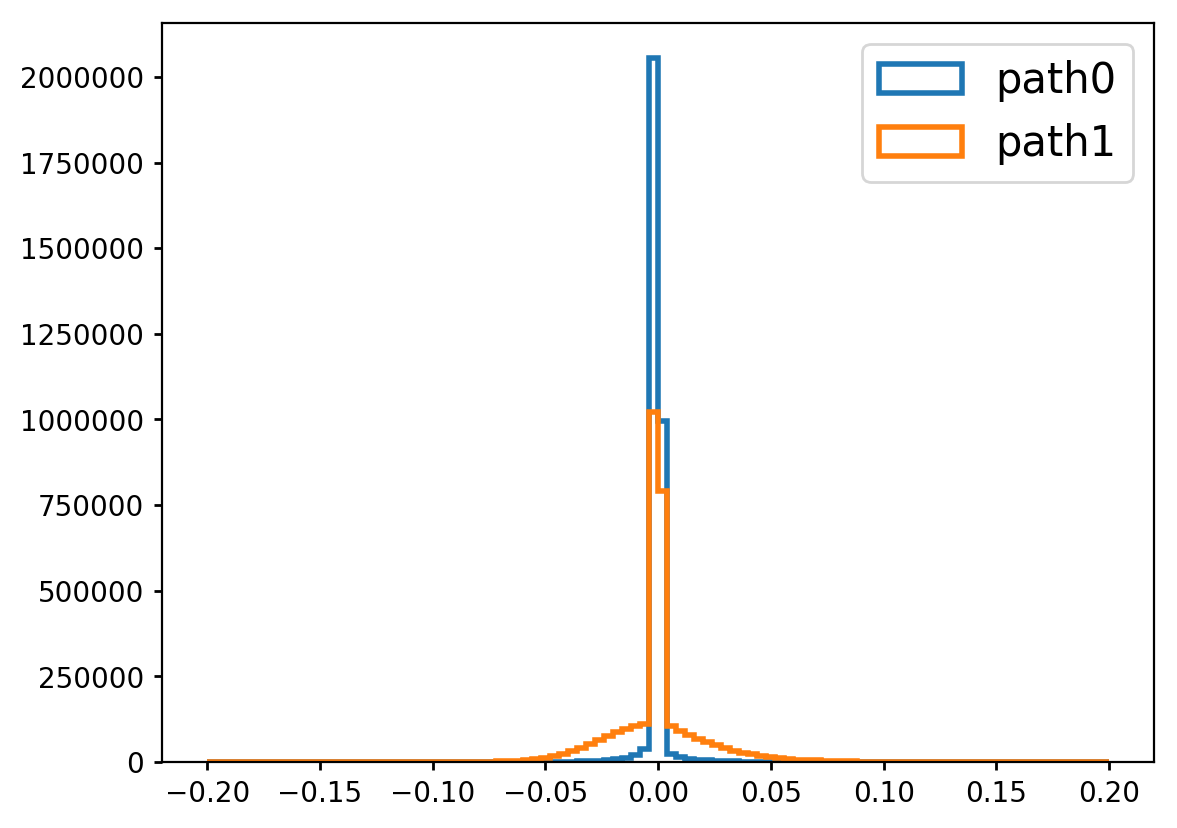}
		\caption{layer 11}
	\end{subfigure}
	\end{center}
	\caption{Weights histograms of parallel convolutional or dense operations at selected layers. Each sub-figure denotes the weight histograms of two parallel operations in the corresponding layer. Differences in histograms in the same layer show that parallel paths have learned different portions of information.}
	\label{fig:weight_hist}
\end{figure*}

One of the reasons for introducing multi-path networks with routing schemes is to group homogeneous feature maps to parallel paths and let the parallel filter sets of the same layer learn different portions of information. Thus, relevant features can be extracted in parallel paths using data-dependent routing during inference. To ensure that our approach enable this, we plot the weights histograms of the VGG13-2 selected layers which carry two parallel convolutions or dense operations on the two sets of feature maps, which are independent of each other. Figure \ref{fig:weight_hist} shows the weights histograms of the two parallel operations at layers 4, 6, 8, and 11. Layer 11 is a dense layer and the other layers are convolutional. The histograms of parallel paths being distinct confirm that the parallel paths have learned different portions of information.

\section{Conclusion}
\label{se:conclusion}

The resource consumption of training deep networks is overwhelming. Thus, designing a network with perfect harmony in depth and width to effectively utilize every trainable parameter is essential. This paper explored ways to strengthen layer-wise feature extraction by implementing parallel paths. In particular, instead of naive network widening, we presented stacking parallel paths into a single network and using novel mechanisms to intelligently route the input among parallel paths end-to-end in a data-dependent manner. Our multi-path networks consistently achieved improved classification accuracy compared to existing widening techniques with similar complexity. Ours also displayed superior performance to existing adaptive learning strategies. Our networks even attained slightly better results than thin deeper networks with similar or more number of parameters. We empirically validated the nature of input dependency of the proposed routing mechanisms and the ability to extract distinct features in parallel paths. Our multi-path networks taking different path combinations according to the input's context is impressive. It can be interpreted as a single adaptive model that softly switches between different sub-modules. Furthermore, it would be intuitive to expand the soft routing towards discrete routing to increase the capacity of the multi-path networks to cater to even multiple datasets at once. 

\section{Data Availability}

CIFAR10 and CIFAR100 datasets \citep{cifar100} are available at \url{https://www.cs.toronto.edu/\~kriz/cifar.html}, and 
ILSVRC 2012 dataset \cite{deng2009imagenet, ILSVRC15} is available at \url{https://www.image-net.org/challenges/LSVRC/2012/}

\section{Declarations}

\textbf{Funding: }
This research is funded by CODEGEN International (Pvt) Ltd, Sri Lanka.
\textbf{Competing Interests: }
The authors have no competing interests to declare that are relevant to the content of this article.

\backmatter

\balance
\bibliography{mybib}


\begin{thebibliography}{61}
\ifx \bisbn   \undefined \def \bisbn  #1{ISBN #1}\fi
\ifx \binits  \undefined \def \binits#1{#1}\fi
\ifx \bauthor  \undefined \def \bauthor#1{#1}\fi
\ifx \batitle  \undefined \def \batitle#1{#1}\fi
\ifx \bjtitle  \undefined \def \bjtitle#1{#1}\fi
\ifx \bvolume  \undefined \def \bvolume#1{\textbf{#1}}\fi
\ifx \byear  \undefined \def \byear#1{#1}\fi
\ifx \bissue  \undefined \def \bissue#1{#1}\fi
\ifx \bfpage  \undefined \def \bfpage#1{#1}\fi
\ifx \blpage  \undefined \def \blpage #1{#1}\fi
\ifx \burl  \undefined \def \burl#1{\textsf{#1}}\fi
\ifx \doiurl  \undefined \def \doiurl#1{\url{https://doi.org/#1}}\fi
\ifx \betal  \undefined \def \betal{\textit{et al.}}\fi
\ifx \binstitute  \undefined \def \binstitute#1{#1}\fi
\ifx \binstitutionaled  \undefined \def \binstitutionaled#1{#1}\fi
\ifx \bctitle  \undefined \def \bctitle#1{#1}\fi
\ifx \beditor  \undefined \def \beditor#1{#1}\fi
\ifx \bpublisher  \undefined \def \bpublisher#1{#1}\fi
\ifx \bbtitle  \undefined \def \bbtitle#1{#1}\fi
\ifx \bedition  \undefined \def \bedition#1{#1}\fi
\ifx \bseriesno  \undefined \def \bseriesno#1{#1}\fi
\ifx \blocation  \undefined \def \blocation#1{#1}\fi
\ifx \bsertitle  \undefined \def \bsertitle#1{#1}\fi
\ifx \bsnm \undefined \def \bsnm#1{#1}\fi
\ifx \bsuffix \undefined \def \bsuffix#1{#1}\fi
\ifx \bparticle \undefined \def \bparticle#1{#1}\fi
\ifx \barticle \undefined \def \barticle#1{#1}\fi
\bibcommenthead
\ifx \bconfdate \undefined \def \bconfdate #1{#1}\fi
\ifx \botherref \undefined \def \botherref #1{#1}\fi
\ifx \url \undefined \def \url#1{\textsf{#1}}\fi
\ifx \bchapter \undefined \def \bchapter#1{#1}\fi
\ifx \bbook \undefined \def \bbook#1{#1}\fi
\ifx \bcomment \undefined \def \bcomment#1{#1}\fi
\ifx \oauthor \undefined \def \oauthor#1{#1}\fi
\ifx \citeauthoryear \undefined \def \citeauthoryear#1{#1}\fi
\ifx \endbibitem  \undefined \def \endbibitem {}\fi
\ifx \bconflocation  \undefined \def \bconflocation#1{#1}\fi
\ifx \arxivurl  \undefined \def \arxivurl#1{\textsf{#1}}\fi
\csname PreBibitemsHook\endcsname

\bibitem{ILSVRC15}
\begin{barticle}
\bauthor{\bsnm{Russakovsky}, \binits{O.}},
\bauthor{\bsnm{Deng}, \binits{J.}},
\bauthor{\bsnm{Su}, \binits{H.}},
\bauthor{\bsnm{Krause}, \binits{J.}},
\bauthor{\bsnm{Satheesh}, \binits{S.}},
\bauthor{\bsnm{Ma}, \binits{S.}},
\bauthor{\bsnm{Huang}, \binits{Z.}},
\bauthor{\bsnm{Karpathy}, \binits{A.}},
\bauthor{\bsnm{Khosla}, \binits{A.}},
\bauthor{\bsnm{Bernstein}, \binits{M.}},
\bauthor{\bsnm{Berg}, \binits{A.C.}},
\bauthor{\bsnm{Fei-Fei}, \binits{L.}}:
\batitle{{ImageNet Large Scale Visual Recognition Challenge}}.
\bjtitle{International Journal of Computer Vision (IJCV)}
\bvolume{115}(\bissue{3}),
\bfpage{211}--\blpage{252}
(\byear{2015}).
\doiurl{10.1007/s11263-015-0816-y}
\end{barticle}
\endbibitem

\bibitem{resnet}
\begin{bchapter}
\bauthor{\bsnm{He}, \binits{K.}},
\bauthor{\bsnm{Zhang}, \binits{X.}},
\bauthor{\bsnm{Ren}, \binits{S.}},
\bauthor{\bsnm{Sun}, \binits{J.}}:
\bctitle{Deep residual learning for image recognition}.
In: \bbtitle{Proceedings of the IEEE Conference on Computer Vision and Pattern
  Recognition (CVPR)},
pp. \bfpage{770}--\blpage{778}
(\byear{2016})
\end{bchapter}
\endbibitem

\bibitem{preact-resnet}
\begin{bchapter}
\bauthor{\bsnm{He}, \binits{K.}},
\bauthor{\bsnm{Zhang}, \binits{X.}},
\bauthor{\bsnm{Ren}, \binits{S.}},
\bauthor{\bsnm{Sun}, \binits{J.}}:
\bctitle{Identity mappings in deep residual networks}.
In: \bbtitle{European Conference on Computer Vision (ECCV)},
pp. \bfpage{630}--\blpage{645}
(\byear{2016}).
\bcomment{Springer}
\end{bchapter}
\endbibitem

\bibitem{fitnet}
\begin{bchapter}
\bauthor{\bsnm{Romero}, \binits{A.}},
\bauthor{\bsnm{Ballas}, \binits{N.}},
\bauthor{\bsnm{Kahou}, \binits{S.E.}},
\bauthor{\bsnm{Chassang}, \binits{A.}},
\bauthor{\bsnm{Gatta}, \binits{C.}},
\bauthor{\bsnm{Bengio}, \binits{Y.}}:
\bctitle{Fitnets: Hints for thin deep nets}.
In: \bbtitle{Proceedings of International Conference on Learning
  Representations (ICLR)}
(\byear{2015})
\end{bchapter}
\endbibitem

\bibitem{vgg}
\begin{botherref}
\oauthor{\bsnm{Simonyan}, \binits{K.}},
\oauthor{\bsnm{Zisserman}, \binits{A.}}:
Very deep convolutional networks for large-scale image recognition.
arXiv preprint arXiv:1409.1556
(2014)
\end{botherref}
\endbibitem

\bibitem{inception}
\begin{bchapter}
\bauthor{\bsnm{Szegedy}, \binits{C.}},
\bauthor{\bsnm{Liu}, \binits{W.}},
\bauthor{\bsnm{Jia}, \binits{Y.}},
\bauthor{\bsnm{Sermanet}, \binits{P.}},
\bauthor{\bsnm{Reed}, \binits{S.}},
\bauthor{\bsnm{Anguelov}, \binits{D.}},
\bauthor{\bsnm{Erhan}, \binits{D.}},
\bauthor{\bsnm{Vanhoucke}, \binits{V.}},
\bauthor{\bsnm{Rabinovich}, \binits{A.}}:
\bctitle{Going deeper with convolutions}.
In: \bbtitle{Proceedings of the IEEE Conference on Computer Vision and Pattern
  Recognition (CVPR)},
pp. \bfpage{1}--\blpage{9}
(\byear{2015})
\end{bchapter}
\endbibitem

\bibitem{wideresnet}
\begin{bchapter}
\bauthor{\bsnm{Zagoruyko}, \binits{S.}},
\bauthor{\bsnm{Komodakis}, \binits{N.}}:
\bctitle{Wide residual networks}.
In: \bbtitle{Proceedings of the British Machine Vision Conference (BMVC)},
pp. \bfpage{87}--\blpage{18712}
(\byear{2016})
\end{bchapter}
\endbibitem

\bibitem{resnetxt}
\begin{bchapter}
\bauthor{\bsnm{Xie}, \binits{S.}},
\bauthor{\bsnm{Girshick}, \binits{R.}},
\bauthor{\bsnm{Doll{\'a}r}, \binits{P.}},
\bauthor{\bsnm{Tu}, \binits{Z.}},
\bauthor{\bsnm{He}, \binits{K.}}:
\bctitle{Aggregated residual transformations for deep neural networks}.
In: \bbtitle{Proceedings of the IEEE Conference on Computer Vision and Pattern
  Recognition (CVPR)},
pp. \bfpage{1492}--\blpage{1500}
(\byear{2017})
\end{bchapter}
\endbibitem

\bibitem{alexnet}
\begin{bchapter}
\bauthor{\bsnm{Krizhevsky}, \binits{A.}},
\bauthor{\bsnm{Sutskever}, \binits{I.}},
\bauthor{\bsnm{Hinton}, \binits{G.E.}}:
\bctitle{Imagenet classification with deep convolutional neural networks}.
In: \bbtitle{Advances in Neural Information Processing Systems},
pp. \bfpage{1097}--\blpage{1105}
(\byear{2012})
\end{bchapter}
\endbibitem

\bibitem{ciregan2012multi}
\begin{bchapter}
\bauthor{\bsnm{Ciregan}, \binits{D.}},
\bauthor{\bsnm{Meier}, \binits{U.}},
\bauthor{\bsnm{Schmidhuber}, \binits{J.}}:
\bctitle{Multi-column deep neural networks for image classification}.
In: \bbtitle{Proceedings of the IEEE Conference on Computer Vision and Pattern
  Recognition (CVPR)},
pp. \bfpage{3642}--\blpage{3649}
(\byear{2012})
\end{bchapter}
\endbibitem

\bibitem{wang2015multi}
\begin{botherref}
\oauthor{\bsnm{Wang}, \binits{M.}}:
Multi-path convolutional neural networks for complex image classification.
arXiv preprint arXiv:1506.04701
(2015)
\end{botherref}
\endbibitem

\bibitem{friedman1991multivariate}
\begin{barticle}
\bauthor{\bsnm{Friedman}, \binits{J.H.}}:
\batitle{Multivariate adaptive regression splines}.
\bjtitle{The annals of statistics}
\bvolume{19}(\bissue{1}),
\bfpage{1}--\blpage{67}
(\byear{1991})
\end{barticle}
\endbibitem

\bibitem{breiman2017classification}
\begin{bbook}
\bauthor{\bsnm{Breiman}, \binits{L.}},
\bauthor{\bsnm{Friedman}, \binits{J.H.}},
\bauthor{\bsnm{Olshen}, \binits{R.A.}},
\bauthor{\bsnm{Stone}, \binits{C.J.}}:
\bbtitle{Classification and Regression Trees}.
\bpublisher{Routledge}, \blocation{???}
(\byear{2017})
\end{bbook}
\endbibitem

\bibitem{jacobs1991adaptive}
\begin{barticle}
\bauthor{\bsnm{Jacobs}, \binits{R.A.}},
\bauthor{\bsnm{Jordan}, \binits{M.I.}},
\bauthor{\bsnm{Nowlan}, \binits{S.J.}},
\bauthor{\bsnm{Hinton}, \binits{G.E.}}:
\batitle{Adaptive mixtures of local experts}.
\bjtitle{Neural computation}
\bvolume{3}(\bissue{1}),
\bfpage{79}--\blpage{87}
(\byear{1991})
\end{barticle}
\endbibitem

\bibitem{jordan1994hierarchical}
\begin{barticle}
\bauthor{\bsnm{Jordan}, \binits{M.I.}},
\bauthor{\bsnm{Jacobs}, \binits{R.A.}}:
\batitle{Hierarchical mixtures of experts and the em algorithm}.
\bjtitle{Neural computation}
\bvolume{6}(\bissue{2}),
\bfpage{181}--\blpage{214}
(\byear{1994})
\end{barticle}
\endbibitem

\bibitem{eigen2013learning}
\begin{botherref}
\oauthor{\bsnm{Eigen}, \binits{D.}},
\oauthor{\bsnm{Ranzato}, \binits{M.}},
\oauthor{\bsnm{Sutskever}, \binits{I.}}:
Learning factored representations in a deep mixture of experts.
arXiv preprint arXiv:1312.4314
(2013)
\end{botherref}
\endbibitem

\bibitem{shazeer2017outrageously}
\begin{botherref}
\oauthor{\bsnm{Shazeer}, \binits{N.}},
\oauthor{\bsnm{Mirhoseini}, \binits{A.}},
\oauthor{\bsnm{Maziarz}, \binits{K.}},
\oauthor{\bsnm{Davis}, \binits{A.}},
\oauthor{\bsnm{Le}, \binits{Q.}},
\oauthor{\bsnm{Hinton}, \binits{G.}},
\oauthor{\bsnm{Dean}, \binits{J.}}:
Outrageously large neural networks: The sparsely-gated mixture-of-experts
  layer.
arXiv preprint arXiv:1701.06538
(2017)
\end{botherref}
\endbibitem

\bibitem{albawi2017understanding}
\begin{bchapter}
\bauthor{\bsnm{Albawi}, \binits{S.}},
\bauthor{\bsnm{Mohammed}, \binits{T.A.}},
\bauthor{\bsnm{Al-Zawi}, \binits{S.}}:
\bctitle{Understanding of a convolutional neural network}.
In: \bbtitle{2017 International Conference on Engineering and Technology
  (ICET)},
pp. \bfpage{1}--\blpage{6}
(\byear{2017}).
\bcomment{Ieee}
\end{bchapter}
\endbibitem

\bibitem{erhan2009visualizing}
\begin{barticle}
\bauthor{\bsnm{Erhan}, \binits{D.}},
\bauthor{\bsnm{Bengio}, \binits{Y.}},
\bauthor{\bsnm{Courville}, \binits{A.}},
\bauthor{\bsnm{Vincent}, \binits{P.}}:
\batitle{Visualizing higher-layer features of a deep network}.
\bjtitle{University of Montreal}
\bvolume{1341}(\bissue{3}),
\bfpage{1}
(\byear{2009})
\end{barticle}
\endbibitem

\bibitem{kahatapitiya2019context}
\begin{bchapter}
\bauthor{\bsnm{Kahatapitiya}, \binits{K.}},
\bauthor{\bsnm{Tissera}, \binits{D.}},
\bauthor{\bsnm{Rodrigo}, \binits{R.}}:
\bctitle{Context-aware automatic occlusion removal}.
In: \bbtitle{2019 IEEE International Conference on Image Processing (ICIP)},
pp. \bfpage{1895}--\blpage{1899}
(\byear{2019}).
\bcomment{IEEE}
\end{bchapter}
\endbibitem

\bibitem{deng2009imagenet}
\begin{bchapter}
\bauthor{\bsnm{Deng}, \binits{J.}},
\bauthor{\bsnm{Dong}, \binits{W.}},
\bauthor{\bsnm{Socher}, \binits{R.}},
\bauthor{\bsnm{Li}, \binits{L.-J.}},
\bauthor{\bsnm{Li}, \binits{K.}},
\bauthor{\bsnm{Fei-Fei}, \binits{L.}}:
\bctitle{Imagenet: A large-scale hierarchical image database}.
In: \bbtitle{2009 IEEE Conference on Computer Vision and Pattern Recognition},
pp. \bfpage{248}--\blpage{255}
(\byear{2009}).
\bcomment{Ieee}
\end{bchapter}
\endbibitem

\bibitem{tissera2019context}
\begin{botherref}
\oauthor{\bsnm{Tissera}, \binits{D.}},
\oauthor{\bsnm{Kahatapitiya}, \binits{K.}},
\oauthor{\bsnm{Wijesinghe}, \binits{R.}},
\oauthor{\bsnm{Fernando}, \binits{S.}},
\oauthor{\bsnm{Rodrigo}, \binits{R.}}:
Context-aware multipath networks.
arXiv preprint arXiv:1907.11519
(2019)
\end{botherref}
\endbibitem

\bibitem{tissera2021feature}
\begin{bchapter}
\bauthor{\bsnm{Tissera}, \binits{D.}},
\bauthor{\bsnm{Vithanage}, \binits{K.}},
\bauthor{\bsnm{Wijesinghe}, \binits{R.}},
\bauthor{\bsnm{Kahatapitiya}, \binits{K.}},
\bauthor{\bsnm{Fernando}, \binits{S.}},
\bauthor{\bsnm{Rodrigo}, \binits{R.}}:
\bctitle{Feature-dependent cross-connections in multi-path neural networks}.
In: \bbtitle{2020 25th International Conference on Pattern Recognition (ICPR)},
pp. \bfpage{4032}--\blpage{4039}
(\byear{2021}).
\bcomment{IEEE}
\end{bchapter}
\endbibitem

\bibitem{lenet}
\begin{barticle}
\bauthor{\bsnm{LeCun}, \binits{Y.}},
\bauthor{\bsnm{Bottou}, \binits{L.}},
\bauthor{\bsnm{Bengio}, \binits{Y.}},
\bauthor{\bsnm{Haffner}, \binits{P.}}:
\batitle{Gradient-based learning applied to document recognition}.
\bjtitle{Proceedings of the IEEE}
\bvolume{86}(\bissue{11}),
\bfpage{2278}--\blpage{2324}
(\byear{1998})
\end{barticle}
\endbibitem

\bibitem{rumelhart1986learning}
\begin{barticle}
\bauthor{\bsnm{Rumelhart}, \binits{D.E.}},
\bauthor{\bsnm{Hinton}, \binits{G.E.}},
\bauthor{\bsnm{Williams}, \binits{R.J.}}:
\batitle{Learning representations by back-propagating errors}.
\bjtitle{Nature}
\bvolume{323},
\bfpage{533}
(\byear{1986})
\end{barticle}
\endbibitem

\bibitem{inception-v4}
\begin{bchapter}
\bauthor{\bsnm{Szegedy}, \binits{C.}},
\bauthor{\bsnm{Ioffe}, \binits{S.}},
\bauthor{\bsnm{Vanhoucke}, \binits{V.}},
\bauthor{\bsnm{Alemi}, \binits{A.A.}}:
\bctitle{Inception-v4, inception-resnet and the impact of residual connections
  on learning}.
In: \bbtitle{AAAI Conference on Artificial Intelligence}
(\byear{2017})
\end{bchapter}
\endbibitem

\bibitem{szegedy2016rethinking}
\begin{bchapter}
\bauthor{\bsnm{Szegedy}, \binits{C.}},
\bauthor{\bsnm{Vanhoucke}, \binits{V.}},
\bauthor{\bsnm{Ioffe}, \binits{S.}},
\bauthor{\bsnm{Shlens}, \binits{J.}},
\bauthor{\bsnm{Wojna}, \binits{Z.}}:
\bctitle{Rethinking the inception architecture for computer vision}.
In: \bbtitle{Proceedings of the IEEE Conference on Computer Vision and Pattern
  Recognition},
pp. \bfpage{2818}--\blpage{2826}
(\byear{2016})
\end{bchapter}
\endbibitem

\bibitem{cross_stich}
\begin{bchapter}
\bauthor{\bsnm{Misra}, \binits{I.}},
\bauthor{\bsnm{Shrivastava}, \binits{A.}},
\bauthor{\bsnm{Gupta}, \binits{A.}},
\bauthor{\bsnm{Hebert}, \binits{M.}}:
\bctitle{Cross-stitch networks for multi-task learning}.
In: \bbtitle{Proceedings of the IEEE Conference on Computer Vision and Pattern
  Recognition (CVPR)},
pp. \bfpage{3994}--\blpage{4003}
(\byear{2016})
\end{bchapter}
\endbibitem

\bibitem{caruana1997multitask}
\begin{barticle}
\bauthor{\bsnm{Caruana}, \binits{R.}}:
\batitle{Multitask learning}.
\bjtitle{Machine learning}
\bvolume{28}(\bissue{1}),
\bfpage{41}--\blpage{75}
(\byear{1997})
\end{barticle}
\endbibitem

\bibitem{thung2018brief}
\begin{barticle}
\bauthor{\bsnm{Thung}, \binits{K.-H.}},
\bauthor{\bsnm{Wee}, \binits{C.-Y.}}:
\batitle{A brief review on multi-task learning}.
\bjtitle{Multimedia Tools and Applications}
\bvolume{77}(\bissue{22}),
\bfpage{29705}--\blpage{29725}
(\byear{2018})
\end{barticle}
\endbibitem

\bibitem{crawshaw2020multi}
\begin{botherref}
\oauthor{\bsnm{Crawshaw}, \binits{M.}}:
Multi-task learning with deep neural networks: A survey.
arXiv preprint arXiv:2009.09796
(2020)
\end{botherref}
\endbibitem

\bibitem{sluice}
\begin{bchapter}
\bauthor{\bsnm{Ruder}, \binits{S.}},
\bauthor{\bsnm{Bingel}, \binits{J.}},
\bauthor{\bsnm{Augenstein}, \binits{I.}},
\bauthor{\bsnm{S{\o}gaard}, \binits{A.}}:
\bctitle{Latent multi-task architecture learning}.
In: \bbtitle{Proceedings of AAAI Conference of Artificial Intelligence},
pp. \bfpage{4822}--\blpage{4829}
(\byear{2019})
\end{bchapter}
\endbibitem

\bibitem{nddr-cnn}
\begin{bchapter}
\bauthor{\bsnm{Gao}, \binits{Y.}},
\bauthor{\bsnm{Ma}, \binits{J.}},
\bauthor{\bsnm{Zhao}, \binits{M.}},
\bauthor{\bsnm{Liu}, \binits{W.}},
\bauthor{\bsnm{Yuille}, \binits{A.L.}}:
\bctitle{Nddr-cnn: Layerwise feature fusing in multi-task cnns by neural
  discriminative dimensionality reduction}.
In: \bbtitle{Proceedings of the IEEE Conference on Computer Vision and Pattern
  Recognition (CVPR)},
pp. \bfpage{3205}--\blpage{3214}
(\byear{2019})
\end{bchapter}
\endbibitem

\bibitem{ha2016hypernetworks}
\begin{bchapter}
\bauthor{\bsnm{Ha}, \binits{D.}},
\bauthor{\bsnm{Dai}, \binits{A.}},
\bauthor{\bsnm{Le}, \binits{Q.V.}}:
\bctitle{Hypernetworks}.
In: \bbtitle{Proceedings of International Conference on Learning
  Representations (ICLR)}
(\byear{2017})
\end{bchapter}
\endbibitem

\bibitem{cai2021dynamic}
\begin{bchapter}
\bauthor{\bsnm{Cai}, \binits{S.}},
\bauthor{\bsnm{Shu}, \binits{Y.}},
\bauthor{\bsnm{Wang}, \binits{W.}}:
\bctitle{Dynamic routing networks}.
In: \bbtitle{Proceedings of the IEEE/CVF Winter Conference on Applications of
  Computer Vision},
pp. \bfpage{3588}--\blpage{3597}
(\byear{2021})
\end{bchapter}
\endbibitem

\bibitem{emrouting}
\begin{bchapter}
\bauthor{\bsnm{Hinton}, \binits{G.E.}},
\bauthor{\bsnm{Sabour}, \binits{S.}},
\bauthor{\bsnm{Frosst}, \binits{N.}}:
\bctitle{Matrix capsules with {EM} routing}.
In: \bbtitle{Proceedings of International Conference on Learning
  Representations (ICLR)}
(\byear{2018})
\end{bchapter}
\endbibitem

\bibitem{hu2018gather}
\begin{bchapter}
\bauthor{\bsnm{Hu}, \binits{J.}},
\bauthor{\bsnm{Shen}, \binits{L.}},
\bauthor{\bsnm{Albanie}, \binits{S.}},
\bauthor{\bsnm{Sun}, \binits{G.}},
\bauthor{\bsnm{Vedaldi}, \binits{A.}}:
\bctitle{Gather-excite: Exploiting feature context in convolutional neural
  networks}.
In: \bbtitle{Advances in Neural Information Processing Systems},
pp. \bfpage{9401}--\blpage{9411}
(\byear{2018})
\end{bchapter}
\endbibitem

\bibitem{hu2017squeeze}
\begin{bchapter}
\bauthor{\bsnm{Hu}, \binits{J.}},
\bauthor{\bsnm{Shen}, \binits{L.}},
\bauthor{\bsnm{Sun}, \binits{G.}}:
\bctitle{Squeeze-and-excitation networks}.
In: \bbtitle{Proceedings of the IEEE Conference on Computer Vision and Pattern
  Recognition (CVPR)},
pp. \bfpage{7132}--\blpage{7141}
(\byear{2018})
\end{bchapter}
\endbibitem

\bibitem{sabour2017dynamic}
\begin{bchapter}
\bauthor{\bsnm{Sabour}, \binits{S.}},
\bauthor{\bsnm{Frosst}, \binits{N.}},
\bauthor{\bsnm{Hinton}, \binits{G.E.}}:
\bctitle{Dynamic routing between capsules}.
In: \bbtitle{Advances in Neural Information Processing Systems},
pp. \bfpage{3856}--\blpage{3866}
(\byear{2017})
\end{bchapter}
\endbibitem

\bibitem{wang2019eca}
\begin{botherref}
\oauthor{\bsnm{Wang}, \binits{Q.}},
\oauthor{\bsnm{Wu}, \binits{B.}},
\oauthor{\bsnm{Zhu}, \binits{P.}},
\oauthor{\bsnm{Li}, \binits{P.}},
\oauthor{\bsnm{Zuo}, \binits{W.}},
\oauthor{\bsnm{Hu}, \binits{Q.}}:
Eca-net: Efficient channel attention for deep convolutional neural networks.
arXiv preprint arXiv:1910.03151
(2019)
\end{botherref}
\endbibitem

\bibitem{convnet-aig}
\begin{bchapter}
\bauthor{\bsnm{Veit}, \binits{A.}},
\bauthor{\bsnm{Belongie}, \binits{S.}}:
\bctitle{Convolutional networks with adaptive inference graphs}.
In: \bbtitle{European Conference on Computer Vision},
pp. \bfpage{3}--\blpage{18}
(\byear{2018})
\end{bchapter}
\endbibitem

\bibitem{blockdrop}
\begin{bchapter}
\bauthor{\bsnm{Wu}, \binits{Z.}},
\bauthor{\bsnm{Nagarajan}, \binits{T.}},
\bauthor{\bsnm{Kumar}, \binits{A.}},
\bauthor{\bsnm{Rennie}, \binits{S.}},
\bauthor{\bsnm{Davis}, \binits{L.S.}},
\bauthor{\bsnm{Grauman}, \binits{K.}},
\bauthor{\bsnm{Feris}, \binits{R.}}:
\bctitle{Blockdrop: Dynamic inference paths in residual networks}.
In: \bbtitle{Proceedings of the IEEE Conference on Computer Vision and Pattern
  Recognition (CVPR)},
pp. \bfpage{8817}--\blpage{8826}
(\byear{2018})
\end{bchapter}
\endbibitem

\bibitem{srivastava2015highway}
\begin{botherref}
\oauthor{\bsnm{Srivastava}, \binits{R.K.}},
\oauthor{\bsnm{Greff}, \binits{K.}},
\oauthor{\bsnm{Schmidhuber}, \binits{J.}}:
Highway networks.
arXiv preprint arXiv:1505.00387
(2015)
\end{botherref}
\endbibitem

\bibitem{rao2018runtime}
\begin{barticle}
\bauthor{\bsnm{Rao}, \binits{Y.}},
\bauthor{\bsnm{Lu}, \binits{J.}},
\bauthor{\bsnm{Lin}, \binits{J.}},
\bauthor{\bsnm{Zhou}, \binits{J.}}:
\batitle{Runtime network routing for efficient image classification}.
\bjtitle{IEEE transactions on pattern analysis and machine intelligence}
\bvolume{41}(\bissue{10}),
\bfpage{2291}--\blpage{2304}
(\byear{2018})
\end{barticle}
\endbibitem

\bibitem{wang2018skipnet}
\begin{bchapter}
\bauthor{\bsnm{Wang}, \binits{X.}},
\bauthor{\bsnm{Yu}, \binits{F.}},
\bauthor{\bsnm{Dou}, \binits{Z.-Y.}},
\bauthor{\bsnm{Darrell}, \binits{T.}},
\bauthor{\bsnm{Gonzalez}, \binits{J.E.}}:
\bctitle{Skipnet: Learning dynamic routing in convolutional networks}.
In: \bbtitle{Proceedings of the European Conference on Computer Vision (ECCV)},
pp. \bfpage{409}--\blpage{424}
(\byear{2018})
\end{bchapter}
\endbibitem

\bibitem{chen2021multipath}
\begin{barticle}
\bauthor{\bsnm{Chen}, \binits{B.}},
\bauthor{\bsnm{Zhao}, \binits{T.}},
\bauthor{\bsnm{Liu}, \binits{J.}},
\bauthor{\bsnm{Lin}, \binits{L.}}:
\batitle{Multipath feature recalibration densenet for image classification}.
\bjtitle{International Journal of Machine Learning and Cybernetics}
\bvolume{12}(\bissue{3}),
\bfpage{651}--\blpage{660}
(\byear{2021})
\end{barticle}
\endbibitem

\bibitem{zhang2022resnest}
\begin{bchapter}
\bauthor{\bsnm{Zhang}, \binits{H.}},
\bauthor{\bsnm{Wu}, \binits{C.}},
\bauthor{\bsnm{Zhang}, \binits{Z.}},
\bauthor{\bsnm{Zhu}, \binits{Y.}},
\bauthor{\bsnm{Lin}, \binits{H.}},
\bauthor{\bsnm{Zhang}, \binits{Z.}},
\bauthor{\bsnm{Sun}, \binits{Y.}},
\bauthor{\bsnm{He}, \binits{T.}},
\bauthor{\bsnm{Mueller}, \binits{J.}},
\bauthor{\bsnm{Manmatha}, \binits{R.}}, \betal:
\bctitle{Resnest: Split-attention networks}.
In: \bbtitle{Proceedings of the IEEE/CVF Conference on Computer Vision and
  Pattern Recognition},
pp. \bfpage{2736}--\blpage{2746}
(\byear{2022})
\end{bchapter}
\endbibitem

\bibitem{yu2021path}
\begin{botherref}
\oauthor{\bsnm{Yu}, \binits{K.}},
\oauthor{\bsnm{Wang}, \binits{X.}},
\oauthor{\bsnm{Dong}, \binits{C.}},
\oauthor{\bsnm{Tang}, \binits{X.}},
\oauthor{\bsnm{Loy}, \binits{C.C.}}:
Path-restore: Learning network path selection for image restoration.
IEEE Transactions on Pattern Analysis and Machine Intelligence
(2021)
\end{botherref}
\endbibitem

\bibitem{huang2017densely}
\begin{bchapter}
\bauthor{\bsnm{Huang}, \binits{G.}},
\bauthor{\bsnm{Liu}, \binits{Z.}},
\bauthor{\bsnm{Van Der~Maaten}, \binits{L.}},
\bauthor{\bsnm{Weinberger}, \binits{K.Q.}}:
\bctitle{Densely connected convolutional networks}.
In: \bbtitle{Proceedings of the IEEE Conference on Computer Vision and Pattern
  Recognition},
pp. \bfpage{4700}--\blpage{4708}
(\byear{2017})
\end{bchapter}
\endbibitem

\bibitem{highway2}
\begin{bchapter}
\bauthor{\bsnm{Srivastava}, \binits{R.K.}},
\bauthor{\bsnm{Greff}, \binits{K.}},
\bauthor{\bsnm{Schmidhuber}, \binits{J.}}:
\bctitle{Training very deep networks}.
In: \bbtitle{Advances in Neural Information Processing Systems},
pp. \bfpage{2377}--\blpage{2385}
(\byear{2015})
\end{bchapter}
\endbibitem

\bibitem{fedus2022review}
\begin{botherref}
\oauthor{\bsnm{Fedus}, \binits{W.}},
\oauthor{\bsnm{Dean}, \binits{J.}},
\oauthor{\bsnm{Zoph}, \binits{B.}}:
A review of sparse expert models in deep learning.
arXiv preprint arXiv:2209.01667
(2022)
\end{botherref}
\endbibitem

\bibitem{chen2022towards}
\begin{botherref}
\oauthor{\bsnm{Chen}, \binits{Z.}},
\oauthor{\bsnm{Deng}, \binits{Y.}},
\oauthor{\bsnm{Wu}, \binits{Y.}},
\oauthor{\bsnm{Gu}, \binits{Q.}},
\oauthor{\bsnm{Li}, \binits{Y.}}:
Towards understanding mixture of experts in deep learning.
arXiv preprint arXiv:2208.02813
(2022)
\end{botherref}
\endbibitem

\bibitem{lepikhin2020gshard}
\begin{botherref}
\oauthor{\bsnm{Lepikhin}, \binits{D.}},
\oauthor{\bsnm{Lee}, \binits{H.}},
\oauthor{\bsnm{Xu}, \binits{Y.}},
\oauthor{\bsnm{Chen}, \binits{D.}},
\oauthor{\bsnm{Firat}, \binits{O.}},
\oauthor{\bsnm{Huang}, \binits{Y.}},
\oauthor{\bsnm{Krikun}, \binits{M.}},
\oauthor{\bsnm{Shazeer}, \binits{N.}},
\oauthor{\bsnm{Chen}, \binits{Z.}}:
Gshard: Scaling giant models with conditional computation and automatic
  sharding.
arXiv preprint arXiv:2006.16668
(2020)
\end{botherref}
\endbibitem

\bibitem{fedus2021switch}
\begin{botherref}
\oauthor{\bsnm{Fedus}, \binits{W.}},
\oauthor{\bsnm{Zoph}, \binits{B.}},
\oauthor{\bsnm{Shazeer}, \binits{N.}}:
Switch transformers: Scaling to trillion parameter models with simple and
  efficient sparsity
(2021)
\end{botherref}
\endbibitem

\bibitem{riquelme2021scaling}
\begin{barticle}
\bauthor{\bsnm{Riquelme}, \binits{C.}},
\bauthor{\bsnm{Puigcerver}, \binits{J.}},
\bauthor{\bsnm{Mustafa}, \binits{B.}},
\bauthor{\bsnm{Neumann}, \binits{M.}},
\bauthor{\bsnm{Jenatton}, \binits{R.}},
\bauthor{\bsnm{Susano~Pinto}, \binits{A.}},
\bauthor{\bsnm{Keysers}, \binits{D.}},
\bauthor{\bsnm{Houlsby}, \binits{N.}}:
\batitle{Scaling vision with sparse mixture of experts}.
\bjtitle{Advances in Neural Information Processing Systems}
\bvolume{34},
\bfpage{8583}--\blpage{8595}
(\byear{2021})
\end{barticle}
\endbibitem

\bibitem{wu2022residual}
\begin{botherref}
\oauthor{\bsnm{Wu}, \binits{L.}},
\oauthor{\bsnm{Liu}, \binits{M.}},
\oauthor{\bsnm{Chen}, \binits{Y.}},
\oauthor{\bsnm{Chen}, \binits{D.}},
\oauthor{\bsnm{Dai}, \binits{X.}},
\oauthor{\bsnm{Yuan}, \binits{L.}}:
Residual mixture of experts.
arXiv preprint arXiv:2204.09636
(2022)
\end{botherref}
\endbibitem

\bibitem{glorot2011deep}
\begin{bchapter}
\bauthor{\bsnm{Glorot}, \binits{X.}},
\bauthor{\bsnm{Bordes}, \binits{A.}},
\bauthor{\bsnm{Bengio}, \binits{Y.}}:
\bctitle{Deep sparse rectifier neural networks}.
In: \bbtitle{Proceedings of the Fourteenth International Conference on
  Artificial Intelligence and Statistics},
pp. \bfpage{315}--\blpage{323}
(\byear{2011})
\end{bchapter}
\endbibitem

\bibitem{cifar100}
\begin{botherref}
\oauthor{\bsnm{Krizhevsky}, \binits{A.}},
\oauthor{\bsnm{Hinton}, \binits{G.}}, et al.:
Learning multiple layers of features from tiny images.
Technical report,
Citeseer
(2009)
\end{botherref}
\endbibitem

\bibitem{ha2016hypernetworks_arxiv}
\begin{botherref}
\oauthor{\bsnm{Ha}, \binits{D.}},
\oauthor{\bsnm{Dai}, \binits{A.}},
\oauthor{\bsnm{Le}, \binits{Q.V.}}:
Hypernetworks.
arXiv preprint arXiv:1609.09106
(2016)
\end{botherref}
\endbibitem

\bibitem{fbresnet}
\begin{botherref}
\oauthor{\bsnm{Facebook}}:
fb.resnet.torch.
Github.
\url{https://github.com/facebookarchive/fb.resnet.torch}
\end{botherref}
\endbibitem

\bibitem{act_max}
\begin{botherref}
\oauthor{\bsnm{Simonyan}, \binits{K.}},
\oauthor{\bsnm{Vedaldi}, \binits{A.}},
\oauthor{\bsnm{Zisserman}, \binits{A.}}:
Deep inside convolutional networks: Visualising image classification models and
  saliency maps.
arXiv preprint arXiv:1312.6034
(2013)
\end{botherref}
\endbibitem

\end{thebibliography}


\end{document}